\documentclass{article}

\usepackage{PRIMEarxiv}

\usepackage[utf8]{inputenc} 
\usepackage[T1]{fontenc}    
\usepackage{hyperref}       
\usepackage{url}            
\usepackage{booktabs}       
\usepackage{amsfonts}       
\usepackage{nicefrac}       
\usepackage{microtype}      
\usepackage{lipsum}
\usepackage{fancyhdr}       
\usepackage{graphicx}       
\graphicspath{{media/}}     
\usepackage{kotex}
\usepackage{amsfonts}       
\usepackage{nicefrac}       
\usepackage{mathtools}

\usepackage{amssymb}
\usepackage{subfig}
\usepackage{tikz}

\usepackage{wrapfig}
\usepackage{algorithm} 
\usepackage{algpseudocode}
\usepackage{multicol}

\title{Memory Association Networks
}

\author{
  *Seokjun Kim$^{1}$, Jaeeun Jang$^{2}$, Yeonju Jang$^{3}$, Seongyune Choi$^{4}$, *Hyeoncheol Kim \\
  Department of Computer Science and Engineering \\
  Korea University \\
  \texttt{\{*melon7607, wkdwodms0779, spring0425, csyun213, *harrykim\}@korea.ac.kr} \\
}

\begin{document}
\maketitle

\begin{abstract}
We introduce memory association networks(MANs) that memorize and remember any data.
This neural network has two memories. One consists of a queue-structured short-term memory to solve the class imbalance problem and long-term memory to store the distribution of objects, introducing the contents of storing and generating various datasets.
\end{abstract}

\keywords{Artificial Association Networks \and The third paper \and More experiments are in progress}

\section{Introduction}
Various networks have been designed in the deep learning field to date. Typically, images, sounds, text, hierarchical, and relational data are learned through the networks, and inductive learning is performed.
But these networks are limited to specific datasets or specific tasks.

Therefore, we designed artificial association networks that can simultaneously learn various datasets in one network like humans.

And in the second study, deductive association networks were proposed to perform deductive reasoning. The main purpose of this study is to use the results of the previous proposition as input to the next proposition, and it was used to solve the compound proposition by learning a simple proposition.
However, it required a lot of input to perform deduction repeatedly. Therefore, we developed a memory device to store inputs and repeatedly recall.

The hippocampus is an organization for humans to remember long-term.
This has been proven by H.M\cite{scoville1957loss, squire2009legacy}. When his hippocampus was removed, previous memories remained, but later information was no longer remembered as long-term memories.

We try to develop a memory device by imitating a human memory device.
First, humans can recall information they have experienced before. And we can choose and recall the memories we want.

And the information is not a single sample and can be modified in various ways whenever recalled. In this paper, this structure is expressed by sampling from the distribution.
We think that the recognized object has a distribution and is similar to the structure of sampling from the distribution.

Therefore, we implemented a long-term memory by combining the conditional variational autoencoder model(CVAE\cite{sohn2015learning}).
CVAE is a generative model by sampling in the distribution with conditions.
We also considered the CGAN\cite{mirza2014conditional} model, but it had the disadvantage of being difficult to learn.

Second, this network should be able to learn various datasets simultaneously and learn in any environment.

The problem is that when the datasets to be learned are diverse, the size of each dataset is different, which may lead to a class imbalance problem.

If 5,000 images, 100,000 sounds, 1,000 text datasets, it could be learned intensively only about sound datasets.

Therefore, we propose a short-term memory to store queues for each class.

This learning method is similar to the replay memory of DQN\cite{mnih2013playing}, and the difference is that queues with maximum size are created for each object class to be stored to solve the class imbalance problem.

And when learning, data samples are put in short-term memory and sampled and learned as much as the batch size.
\section{Related works}
\label{sec:related_works}

\paragraph{Artificial Association Networks\cite{kim2021association}}
This theory is for integrating all existing networks and developing them into new neural networks.
Therefore, various datasets can be processed at the same time, and various models are integrated, and it is a domain general learning method.

In this study, instead of using a fixed sequence layer, this network defines a data structure called an neuro tree and learns according to the structure of the tree.
The data structure defined $\mathbf{AN} = \{x, t, \mathbf{A}_c, \mathbf{C}\}$, which is designed to store various types of data in memory using this model.
And propagation method is conducted using recursive convolution called depth first convolution(DFC) and depth first deconvolution(DFD).

\paragraph{Deductive Association Networks\cite{kim2021deductive}}
We proposed deductive association networks to solve compound propositions by learning simple propositions.
We defined the task by the group theory of modern algebra, and it showed that the compound proposition is well performed even if the deduction is performed several times.

\paragraph{Conditional Variational Autoencoders}
This study is a generative model and we designed a long-term memory using the conditional variational autoencoder\cite{sohn2015learning} model.
Condition information is used at the beginning of the encoder and decoder parts.
When the desired conditions are entered into the decoder, a new sample is generated.

\paragraph{Replay Memory}
Replay memory is a methodology used in DQN\cite{mnih2013playing}.
Replay memory stored samples appearing in the reinforcement learning environment in the queue structure, and then sampled and learned mini-batch-samples in memory, thereby stabilizing learning in the network.
We designed a short-term memory by slightly modifying this process.

\section{Model : Memory Association Networks}
\begin{figure}[h]
\centering
{\includegraphics[width=0.70\textwidth]{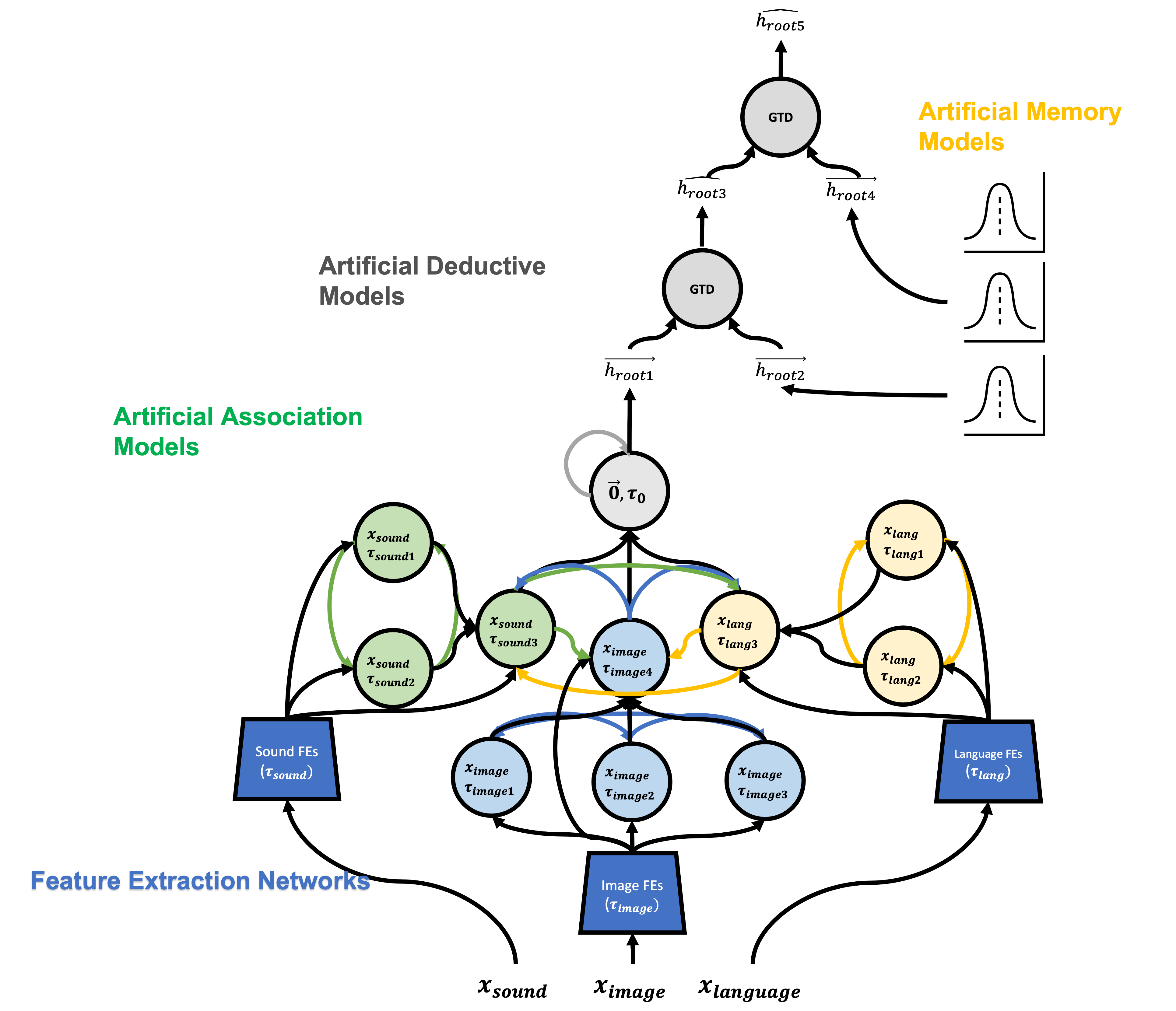}}
\caption{ Ideal Memory Modeling }
\label{fig:long-term}
\vspace{-10px}
\end{figure}
Our perspective on proposing a memory network is as follows.
First, when information is collected from various sensory organs in the association area, the information is transmitted to the frontal lobe.
The frontal lobe controls impulsive behavior or performs optimal behavior for a situation; it controls based on information obtained from past experiences or actions.

If so, it means that the frontal lobe needs a memory device to work, and the frontal lobe and memory have a profound correlation. 
In this way, we need to input every step to perform deduction, and if we have a memory device, we can use the previously inputted information(previous-root-vectors) as experiences.
\subsection{Short-term Memory}
\label{sec:short-term}
\begin{figure}[h]
\centering
{\includegraphics[width=0.70\textwidth]{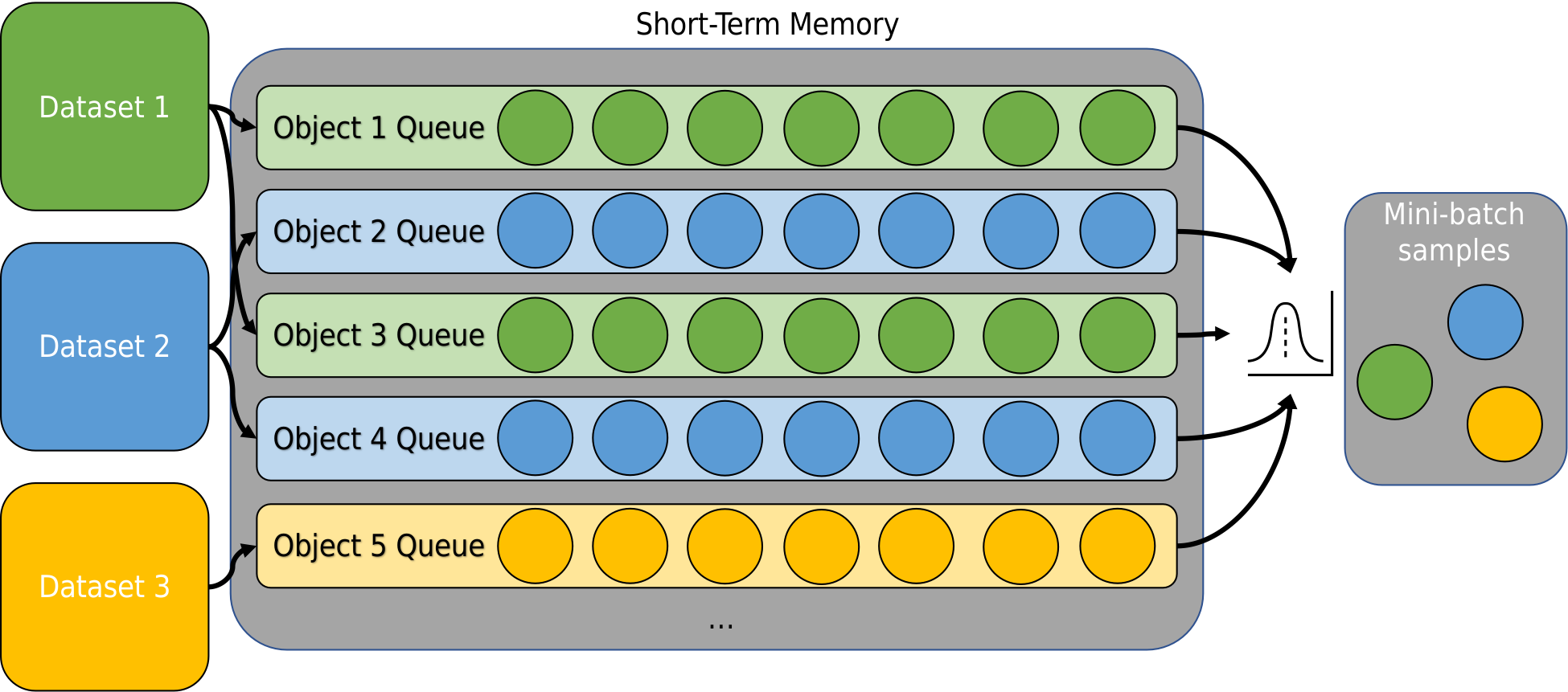}}
\caption{ Short-term Memory }
\label{fig:short-term}
\vspace{-10px}
\end{figure}
Short-term memory exists because learning various datasets can cause different sizes of each dataset and cause class imbalance problems.
The short-term memory generates several queues with a maximum size for each object(class) to be stored.
Random sampling is performed to solve the class imbalance problem when samples of each object are appropriately stored.

And each queue has an object number, and we convert this number into one-hot-vector and use it as key vectors and labels for sampling the object.

\begin{equation}
object_{ij} \in Queue_{i} \subseteq memory_{short}
\label{eqn:short-term-memory}
\end{equation}
$i$ is object number, It can be used as a label for supervised learning.

The importance of short-term memory is to learn in any environment.
When we generally learn, we distribute the number of data fairly by class, but this is not realistic.

A short-term memory device is to make the input information the same size.

\subsection{Long-term Memory (Type A)}

\begin{figure}[h]
\centering
{\includegraphics[width=0.70\textwidth]{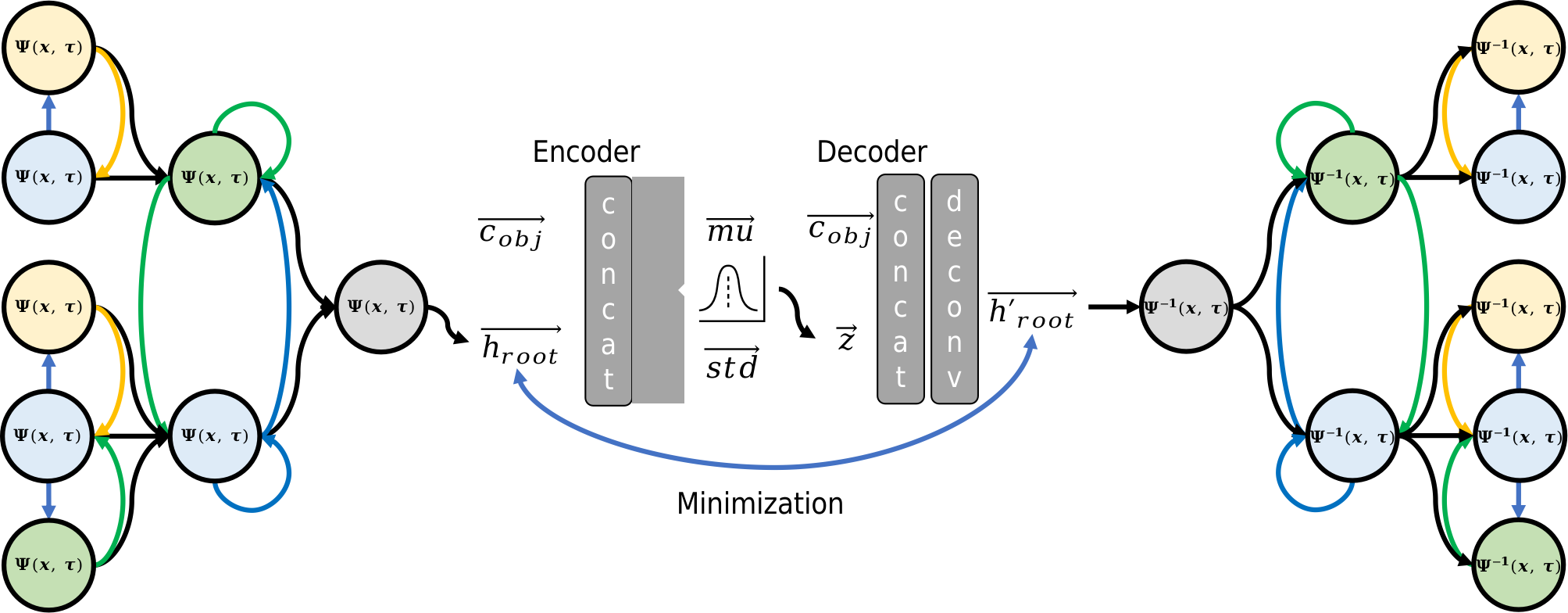}}
\caption{ Long-term Memory (CVAE model)}
\label{fig:long-term}
\vspace{-10px}
\end{figure}

This network uses the existing conditional vae model. And artificial association models are used for the encoder and decoder parts, and one additional loss term is added to restore the root vector.
The purpose of this memory network is "to produce the root vector" again by the object number.
Therefore, the condition information is input with the root vector to the encoder part and it is input of the decoder part with the generated $\overrightarrow{z}$.

We revised the model slightly because we thought it would be more simplifiable. The modified model is introduced in the next section. ($\epsilon \sim N(0,\mathbf{I})$ )
\begin{equation}
\overrightarrow{h}_{ij,root} = DFC(AT_{ij}, 0)
\label{eqn:cvae-long-term}
\end{equation}
\begin{equation}
encoder_{long}(\overrightarrow{c}_{i},\overrightarrow{h}_{ij,root}) = mu_{ij}, std_{ij}
\label{eqn:cvae-long-term}
\end{equation}
\begin{equation}
\overrightarrow{z}_{ij} = mu_{ij} + std^{2}_{ij} \circ \epsilon
\label{eqn:cvae-long-term}
\end{equation}
\begin{equation}
decoder_{long}(\overrightarrow{c}_{i},\overrightarrow{z}_{ij}) = \hat{h}_{ij,root}
\label{eqn:cvae-long-term}
\end{equation}
\begin{equation}
AT_{ij} = DFD(\hat{h}_{ij,root},AT_{ij})
\label{eqn:cvae-long-term}
\end{equation}
\begin{equation}
loss_{reconstruction} = reconstruction_{recursive}(AT.x_{ij}, AT.\hat{x}_{ij})
\label{eqn:cvae-long-term}
\end{equation}

\begin{equation}
Q_{ij} = N(mu_{ij}, std_{ij}), P = N(0, \mathbf{I})
\label{eqn:cvae-long-term}
\end{equation}
\begin{equation}
loss = D_{KL}(Q_{ij}||P) - loss_{reconstruction} + \lambda MSE(\overrightarrow{h}_{ij,root}, \hat{h}_{ij,root})
\label{eqn:cvae-long-term}
\end{equation}

\subsection{Long-term Memory (Type B)}
\begin{figure}[h]
\centering
{\includegraphics[width=0.70\textwidth]{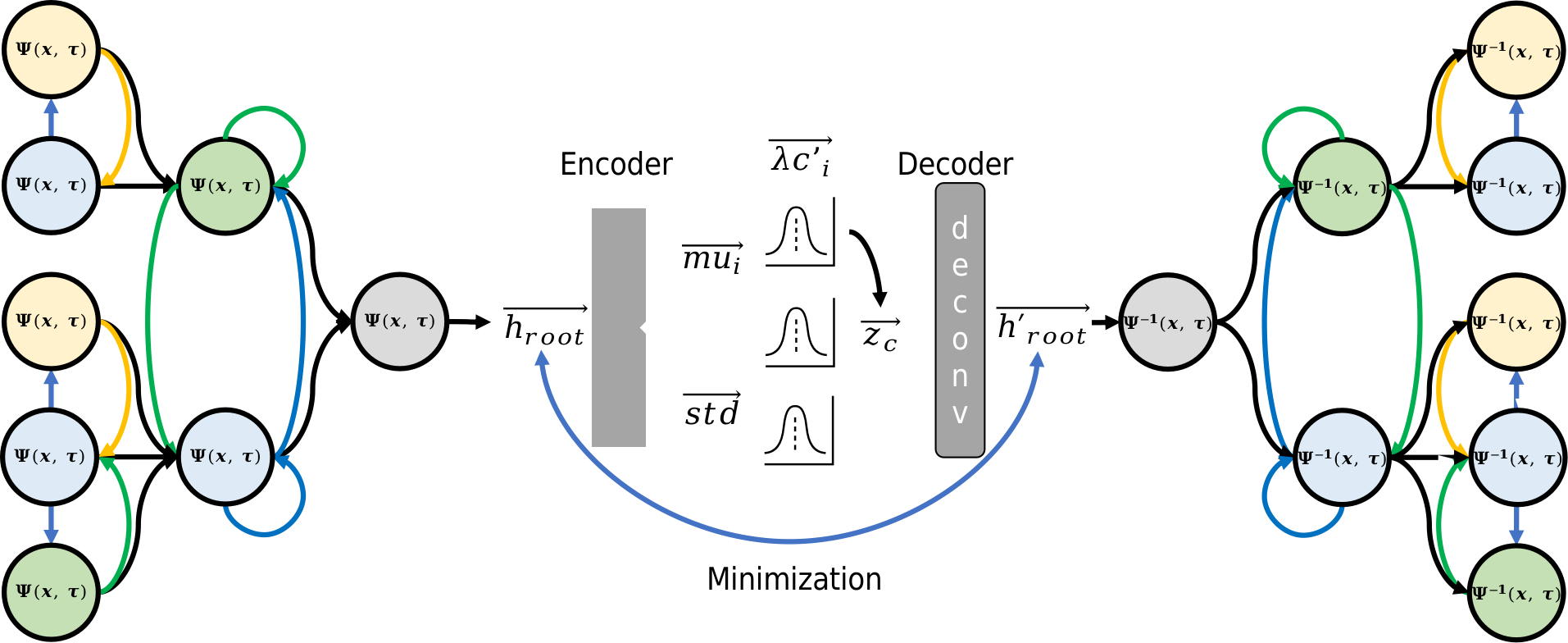}}
\caption{ Long-term Memory (Recognition + Memorization)}
\label{fig:long-term}
\vspace{-10px}
\end{figure}
The second model simultaneously learns recognition learning and memorization.
We modified vae model\cite{kingma2013auto} and trained it to perform the recognition process together.
We converted the object number($i$) of each object into one hot vector($\overrightarrow{c}_{i}$) and approximated it as a distribution $~N(c, I)$
In other words, this model approximates different distributions for each class of the object.
The characteristic of this model is that the $mu$ of the distribution means the class vectors, so it contains information on the class($mu_{i}$).

However, if one-hot vector is used as $mu_{i}$, and std is $I$, a part that overlaps with the distribution of other objects occurs.

\begin{figure}[h]
\centering
\subfloat[onehot]{
{\includegraphics[width=4.5cm]{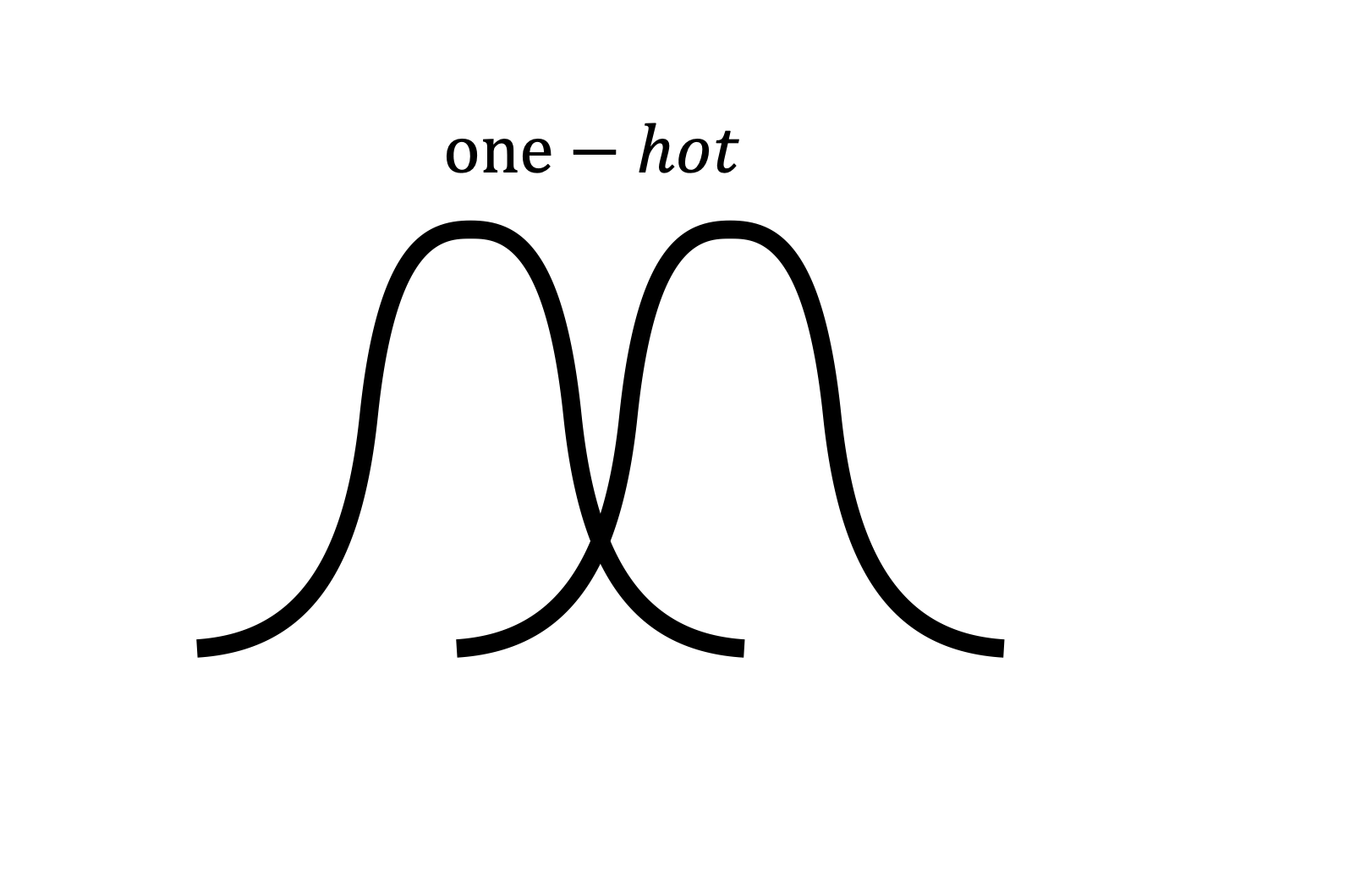}}
}
\subfloat[lambda hot]{
{\includegraphics[width=4.5cm]{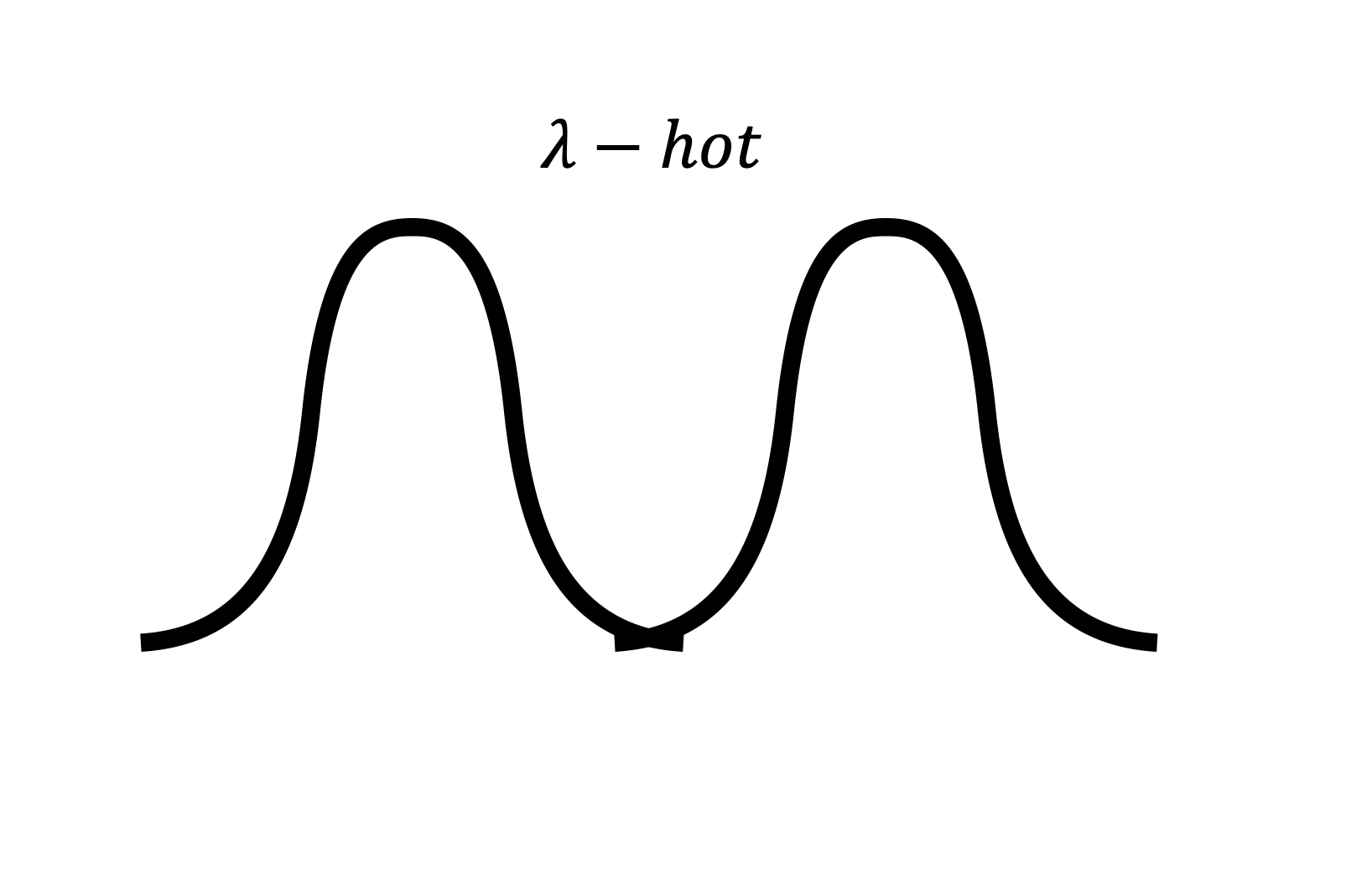}}
}
\caption{overlap}
\label{fig:deductive_reasoning}
\end{figure}

\begin{equation}
\overrightarrow{c}_{i} = onehot(i)
\label{eqn:cvae-long-term}
\end{equation}

So we create a $\lambda$-hot vector. It multiplies the lambda($\lambda$) to a one-hot vector.
The purpose of the lambda value is to ensure that the distributions do not overlap each other.
Consider the standard deviation value of the approximate distribution so that the two distributions intersect at the point where the z-score value becomes 3.

\begin{equation}
\lambda \overrightarrow{c}_{i} = \lambda hot(i)
\label{eqn:cvae-long-term}
\end{equation}
Therefore the standard deviation value is 1, the two distributions must be at a distance of 3($\sigma \times z_{score}$) + 3($\sigma \times z_{score}$) = 6 or more.
Therefore, since the $\lambda$ value is 6 or more, it should be learned as [$\lambda$, 0, 0, ...] instead of [1, 0, 0, ...]. ($N(\lambda c, \mathbf{I})$)

And This model has one problem that if the dimension of the $\lambda$-hot vector is small, too much information is lost.
Therefore, we can increase the dimension of the $\lambda$-hot vector by zero padding to the $\lambda$-hot vector. 
\begin{equation}
\lambda \overrightarrow{c}'_{i} = [\lambda hot(i),\overrightarrow{0}]
\label{eqn:cvae-long-term}
\end{equation}

This process, when combined with our other study, the deductive model, can be replaced by this method without performing a recognition task.
Because, when we learn with a class vector using kl divergence, we generally learn using a standard normal distribution as a model.
\begin{equation}
Q_{ij} = N(mu_{ij}, std_{ij}), P_{i} = N(\lambda_{1}\overrightarrow{c}'_{i}, \mathbf{I})
\label{eqn:cvae-long-term}
\end{equation}
\begin{equation}
loss = D_{KL}(Q_{ij}||P_{i}) - loss_{reconstruction} + \lambda_{2} MSE(\overrightarrow{h}_{ij,root}, \hat{h}_{ij,root})
\label{eqn:cvae-long-term}
\end{equation}

\begin{equation}
distributions_{i} \in memory_{long}
\label{eqn:long-term-memory}
\end{equation}

\begin{equation}
f(memory_{short}) \to distributions \to memory_{long}
\label{eqn:short-term-memory}
\end{equation}
The queue of each object becomes data samples for generating the distribution of the object.

\begin{equation}
(memory_{short} \cup memory_{long}) \subseteq memory
\label{eqn:short-term-memory}
\end{equation}
These two memories are subsets of a set called memory, and other storage devices can be added later.

\section{Experimental Results}

\subsection{MNIST Dataset}

\begin{figure}[h]
\centering
\begin{tikzpicture}[picture format/.style={inner sep=0pt,}]
  \node[picture format]                   (A1)               
  {\includegraphics[width=0.10\textwidth]{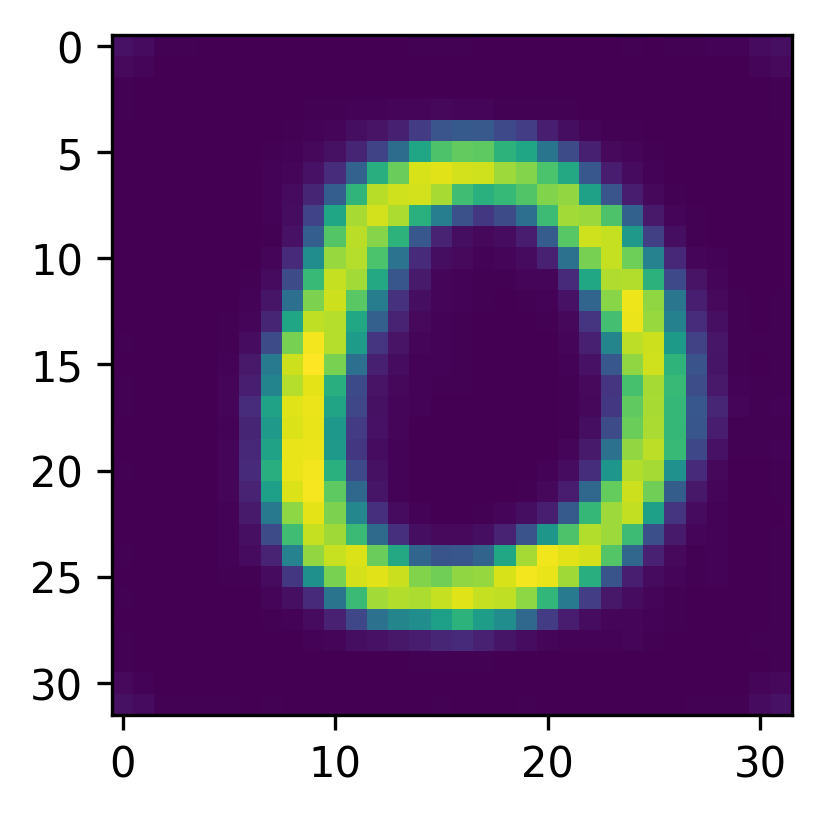}};
  \node[picture format,anchor=north]      (B1) at (A1.south)  {\includegraphics[width=0.10\textwidth]{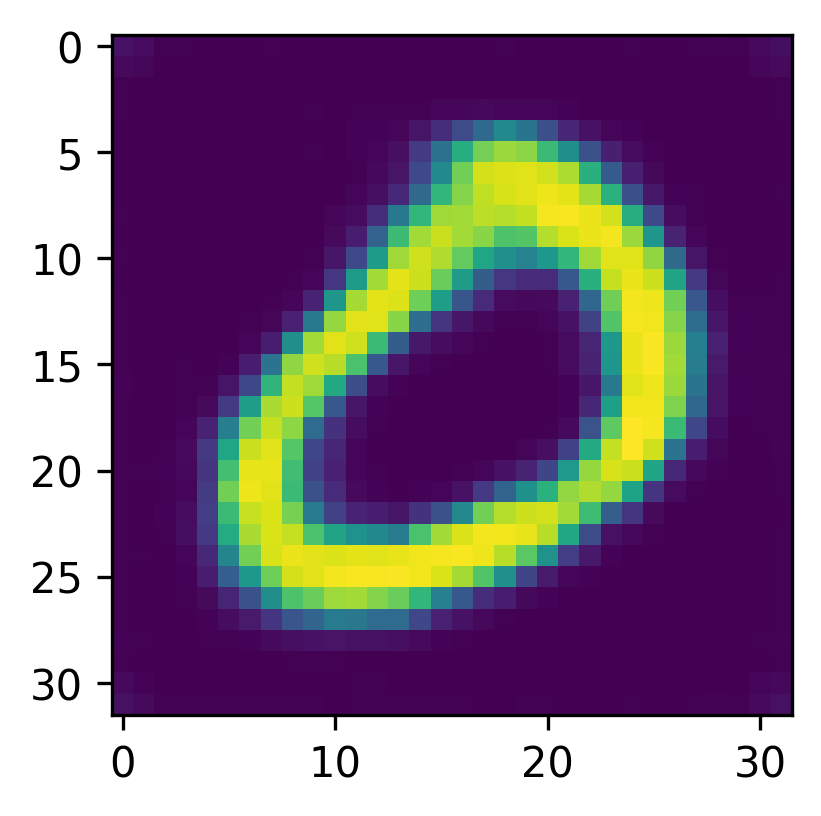}};
  \node[picture format,anchor=north]      (C1) at (B1.south) {\includegraphics[width=0.10\textwidth]{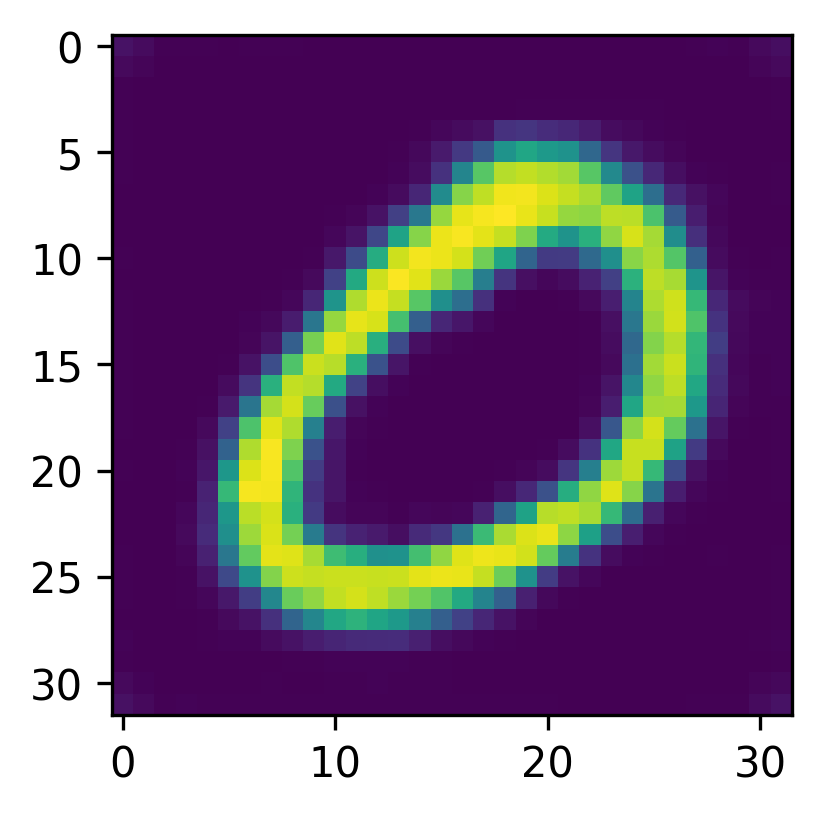}};  
  \node[picture format,anchor=north]      (D1) at (C1.south) {\includegraphics[width=0.10\textwidth]{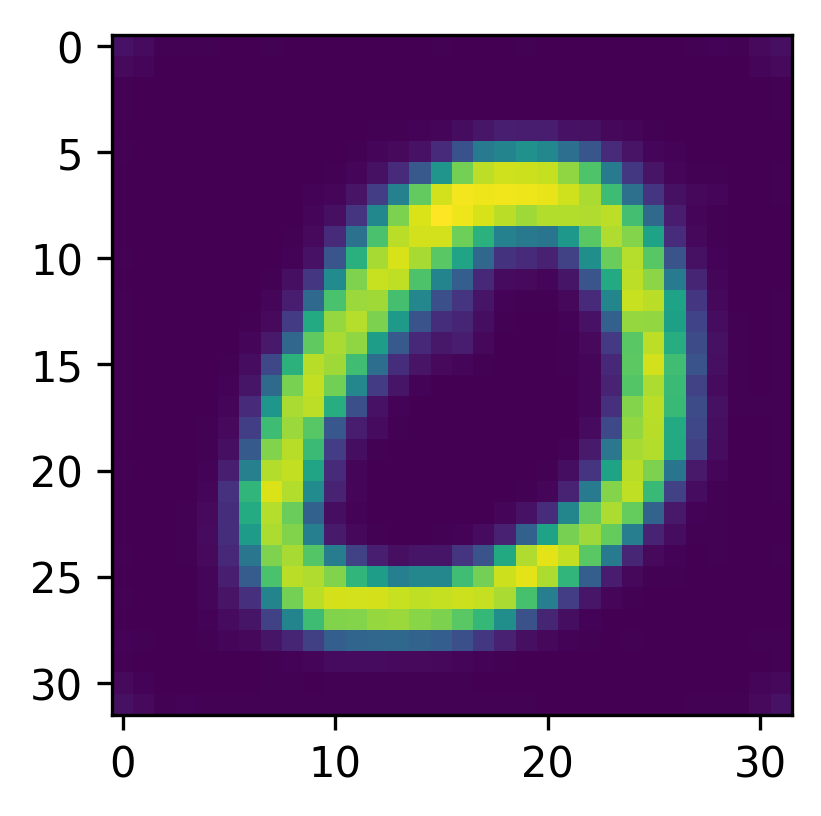}};

  \node[picture format,anchor=north west] (A2) at (A1.north east) {\includegraphics[width=0.10\textwidth]{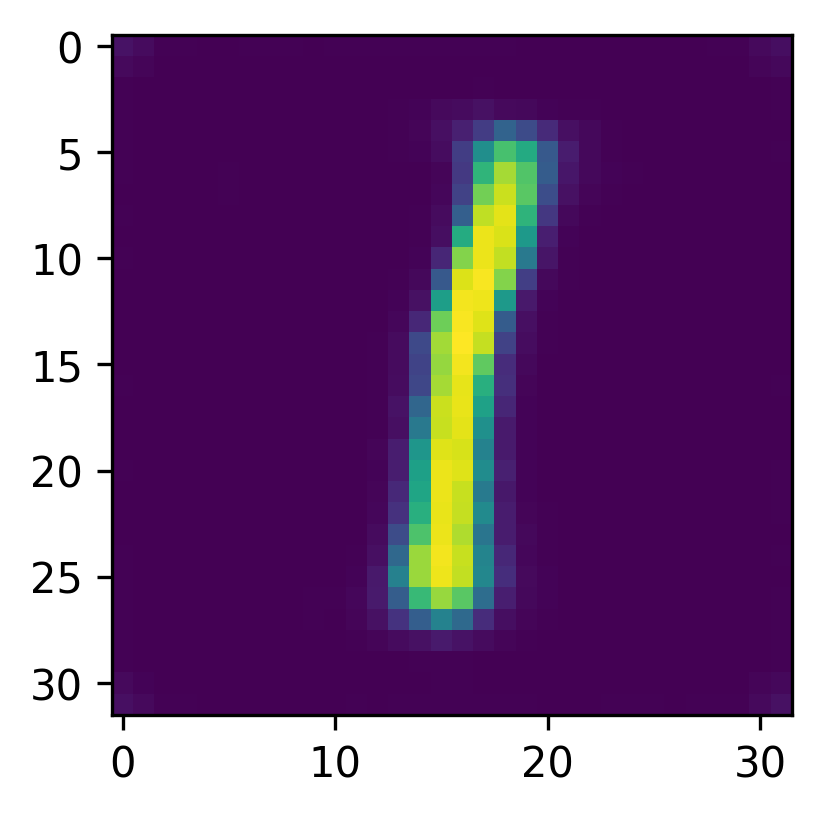}};
  \node[picture format,anchor=north]      (B2) at (A2.south)      {\includegraphics[width=0.10\textwidth]{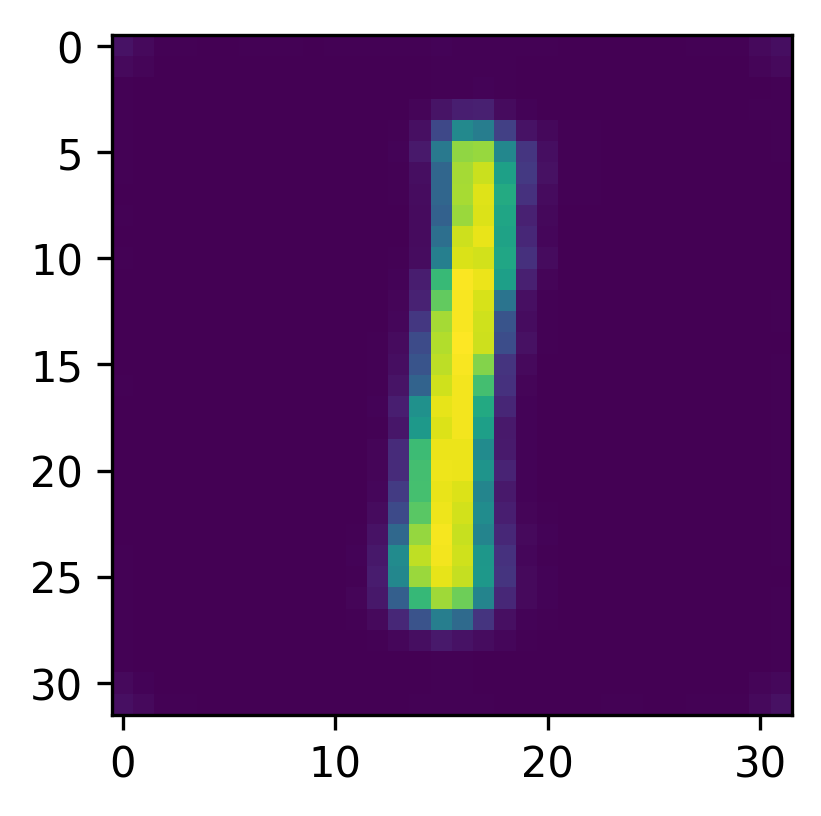}};
  \node[picture format,anchor=north]      (C2) at (B2.south)      {\includegraphics[width=0.10\textwidth]{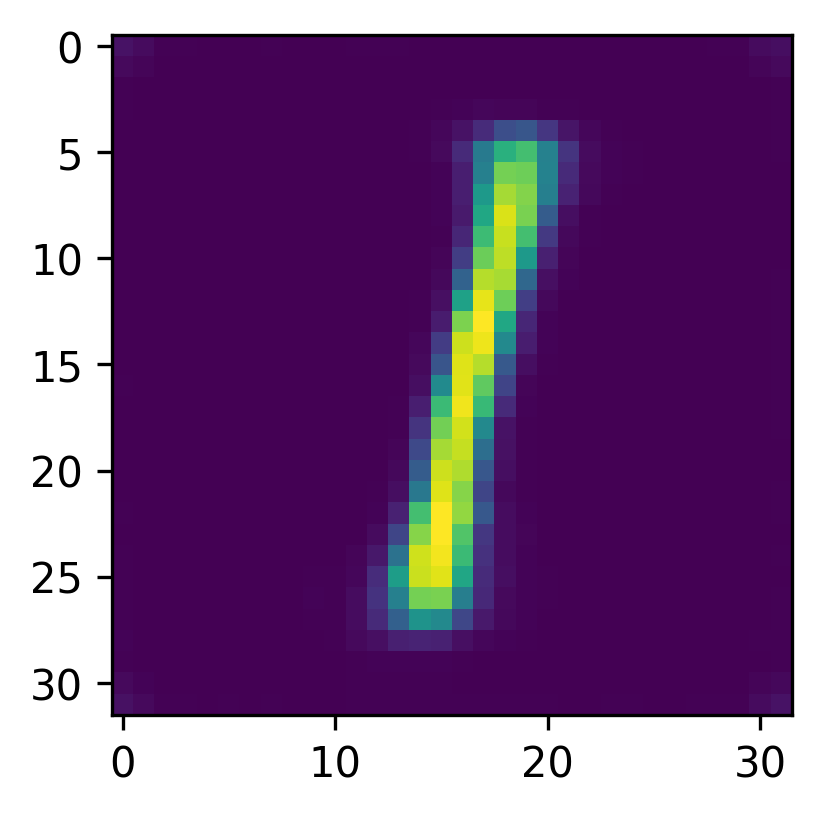}};
  \node[picture format,anchor=north]      (D2) at (C2.south)      {\includegraphics[width=0.10\textwidth]{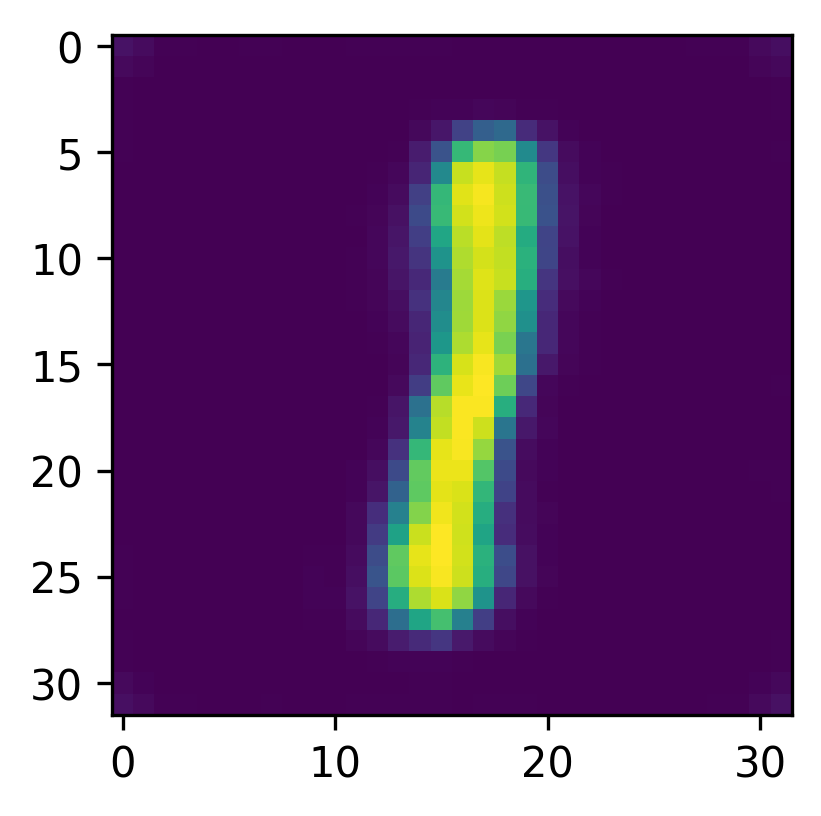}};

  \node[picture format,anchor=north west] (A3) at (A2.north east) {\includegraphics[width=0.10\textwidth]{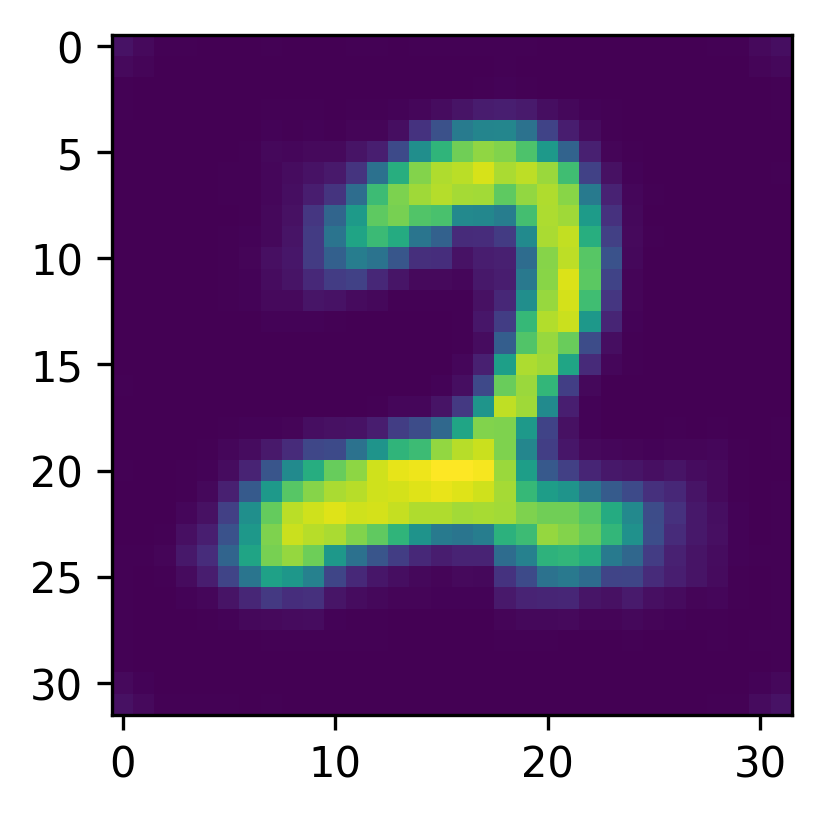}};
  \node[picture format,anchor=north]      (B3) at (A3.south)      {\includegraphics[width=0.10\textwidth]{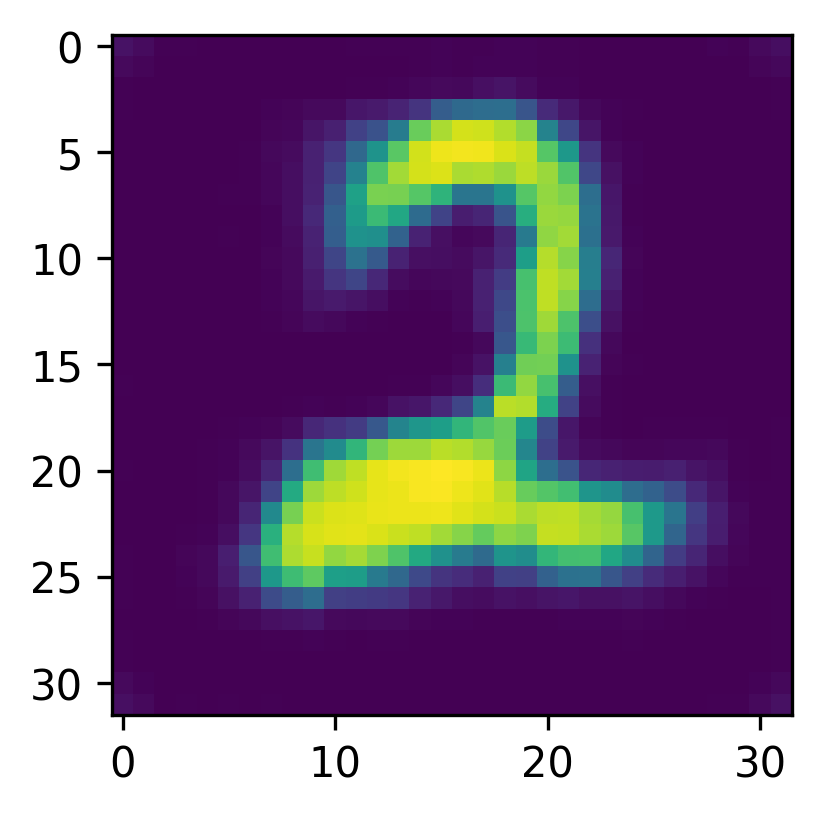}};
  \node[picture format,anchor=north]      (C3) at (B3.south)      {\includegraphics[width=0.10\textwidth]{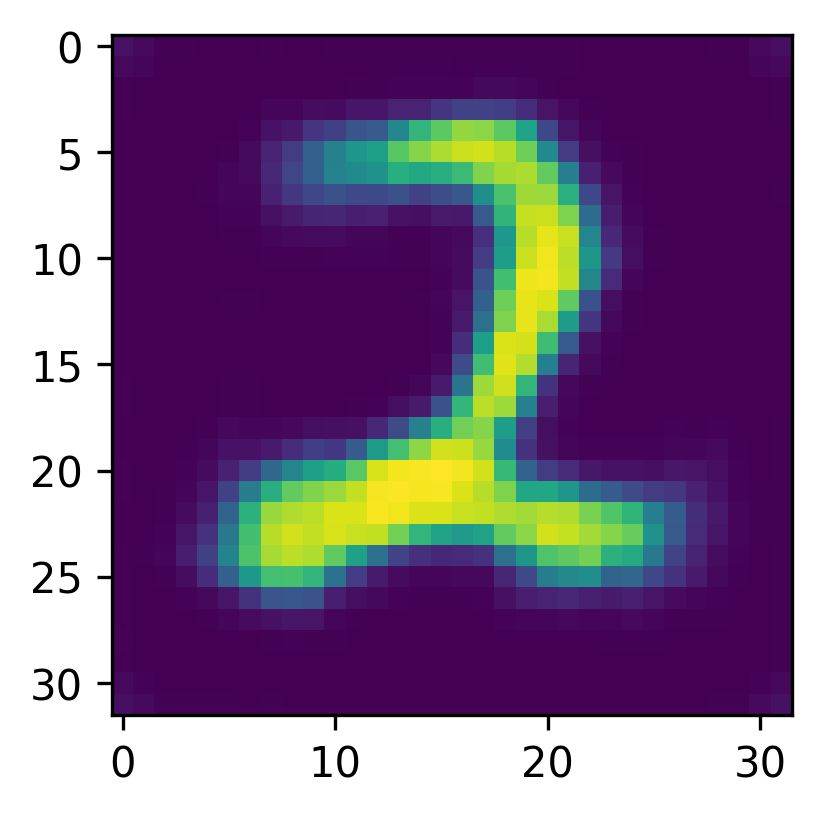}};
  \node[picture format,anchor=north]      (D3) at (C3.south)      {\includegraphics[width=0.10\textwidth]{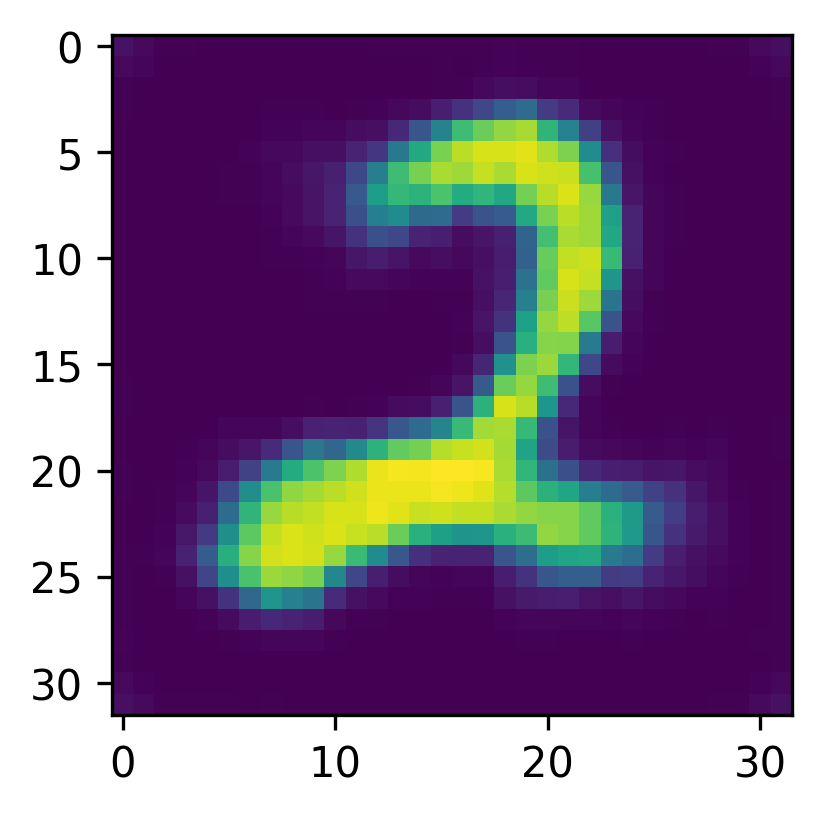}};

  \node[picture format,anchor=north west] (A4) at (A3.north east) {\includegraphics[width=0.10\textwidth]{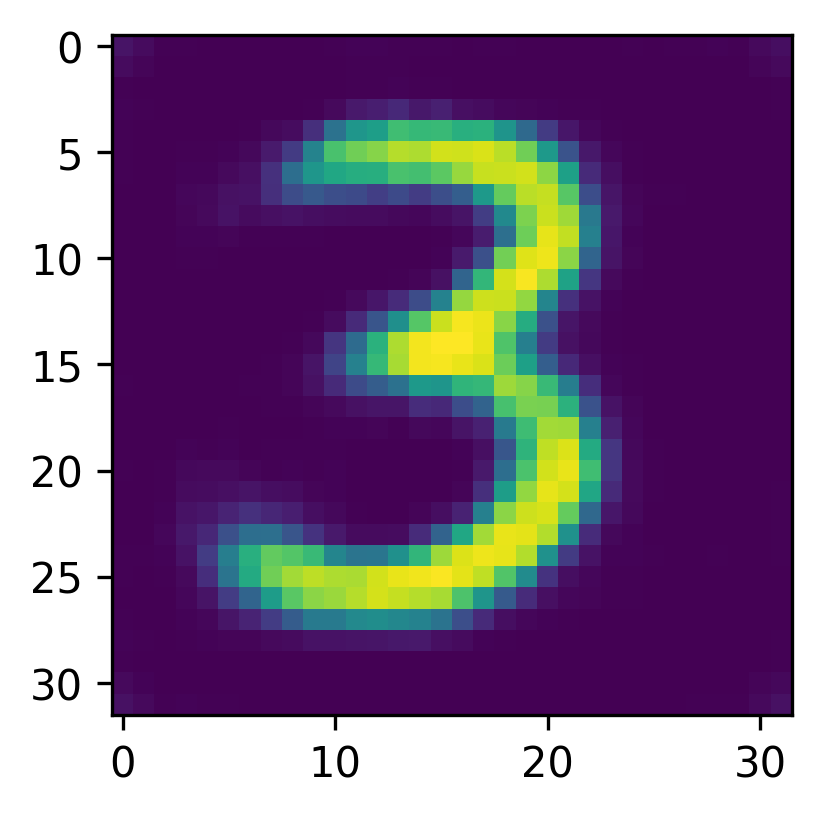}};
  \node[picture format,anchor=north]      (B4) at (A4.south)      {\includegraphics[width=0.10\textwidth]{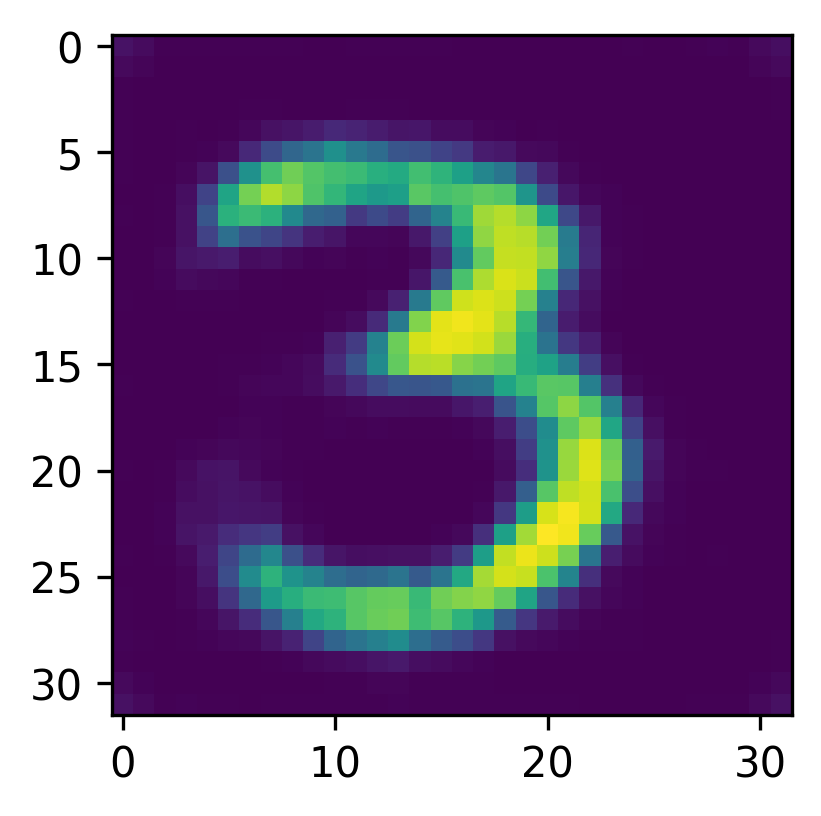}};
  \node[picture format,anchor=north]      (C4) at (B4.south)      {\includegraphics[width=0.10\textwidth]{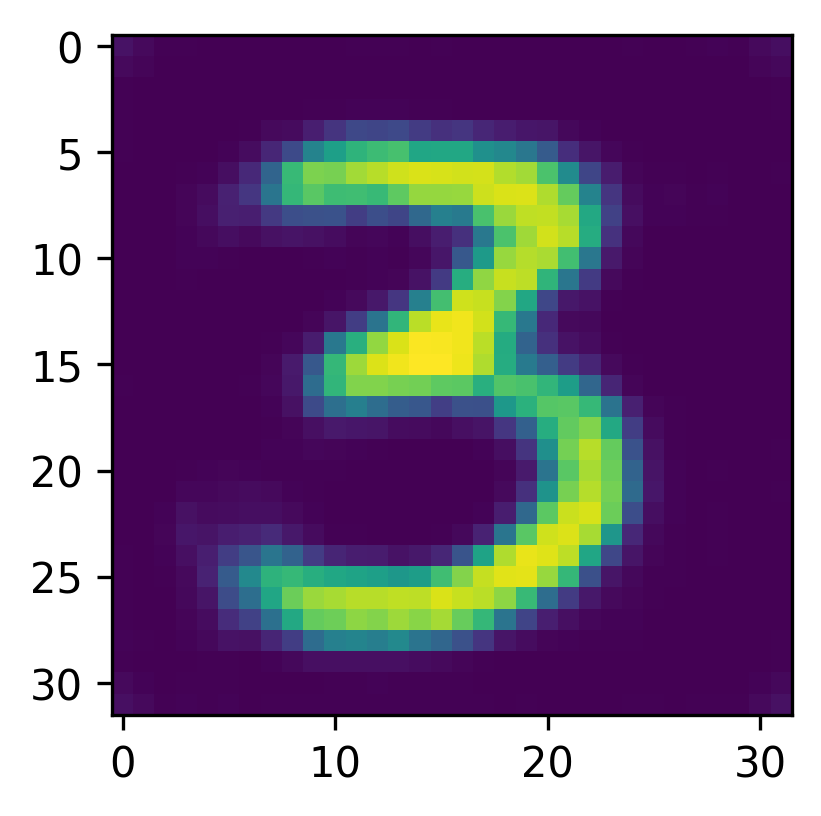}};
  \node[picture format,anchor=north]      (D4) at (C4.south)      {\includegraphics[width=0.10\textwidth]{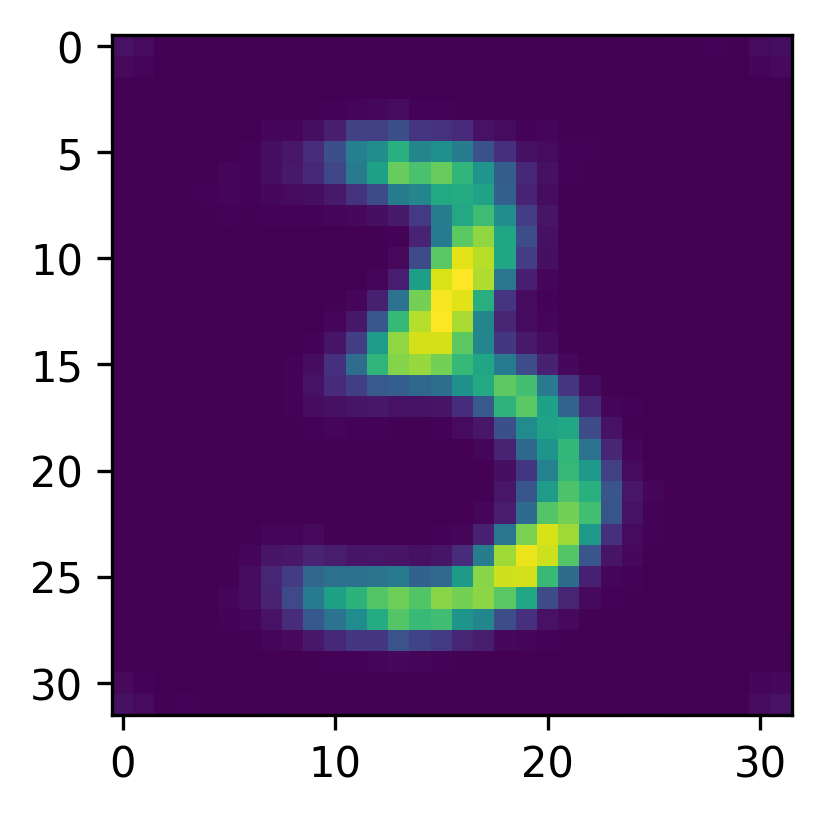}};

  \node[picture format,anchor=north west] (A5) at (A4.north east) {\includegraphics[width=0.10\textwidth]{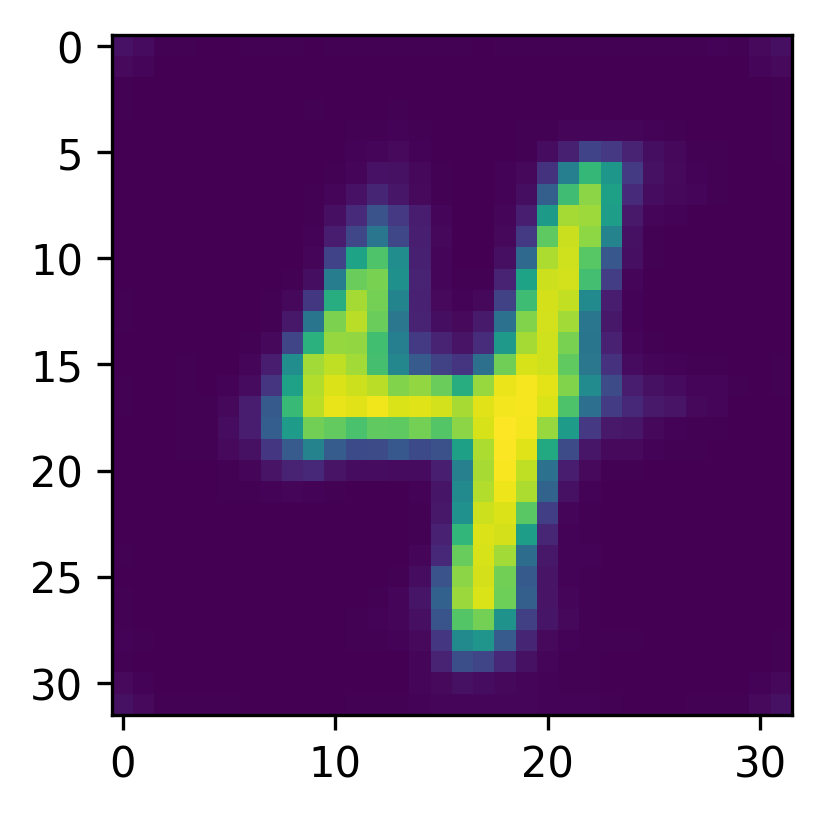}};
  \node[picture format,anchor=north]      (B5) at (A5.south)      {\includegraphics[width=0.10\textwidth]{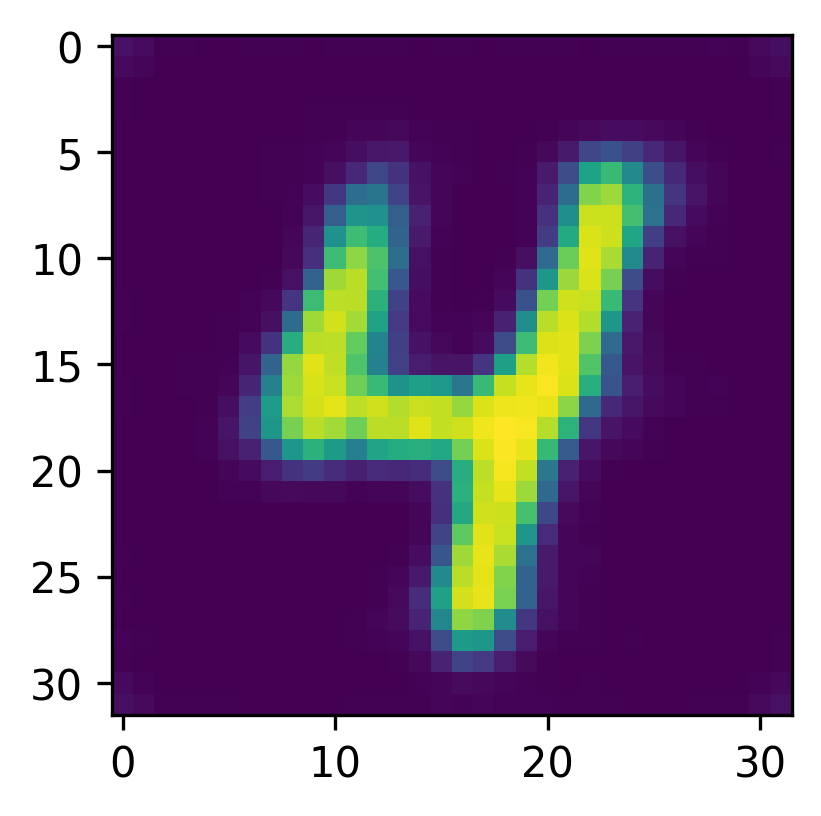}};
  \node[picture format,anchor=north]      (C5) at (B5.south)      {\includegraphics[width=0.10\textwidth]{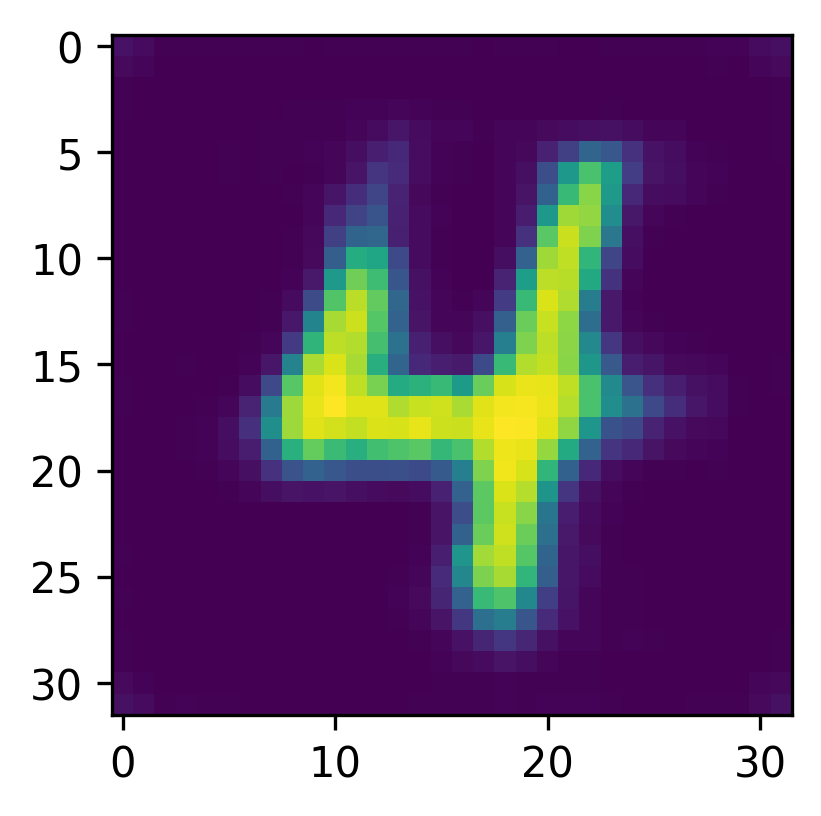}};
  \node[picture format,anchor=north]      (D5) at (C5.south)      {\includegraphics[width=0.10\textwidth]{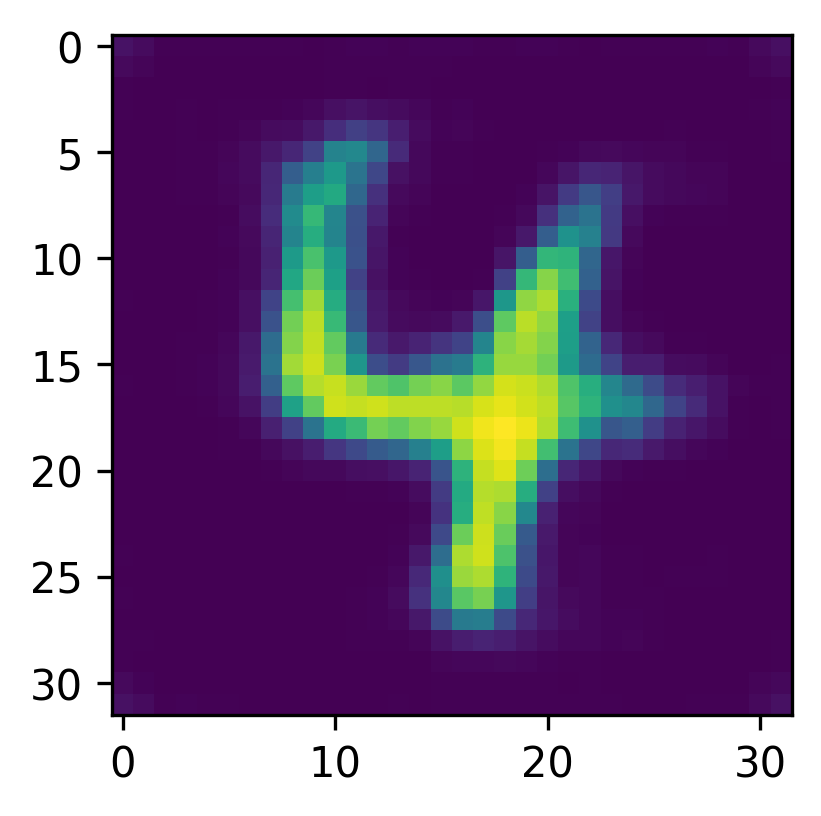}};

  \node[picture format,anchor=north west] (A6) at (A5.north east) {\includegraphics[width=0.10\textwidth]{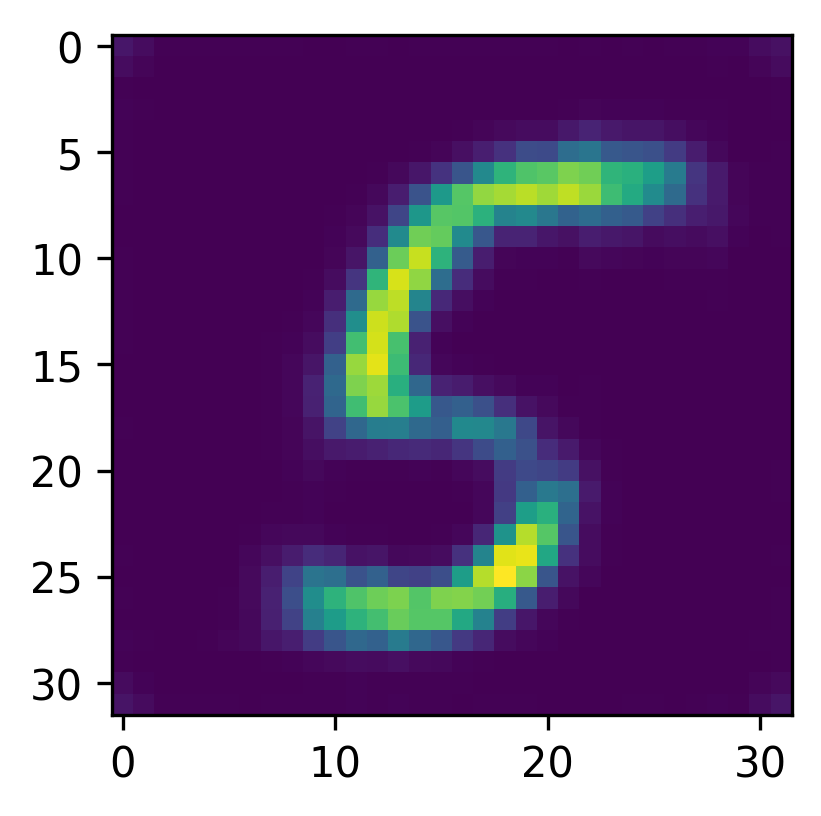}};
  \node[picture format,anchor=north]      (B6) at (A6.south)      {\includegraphics[width=0.10\textwidth]{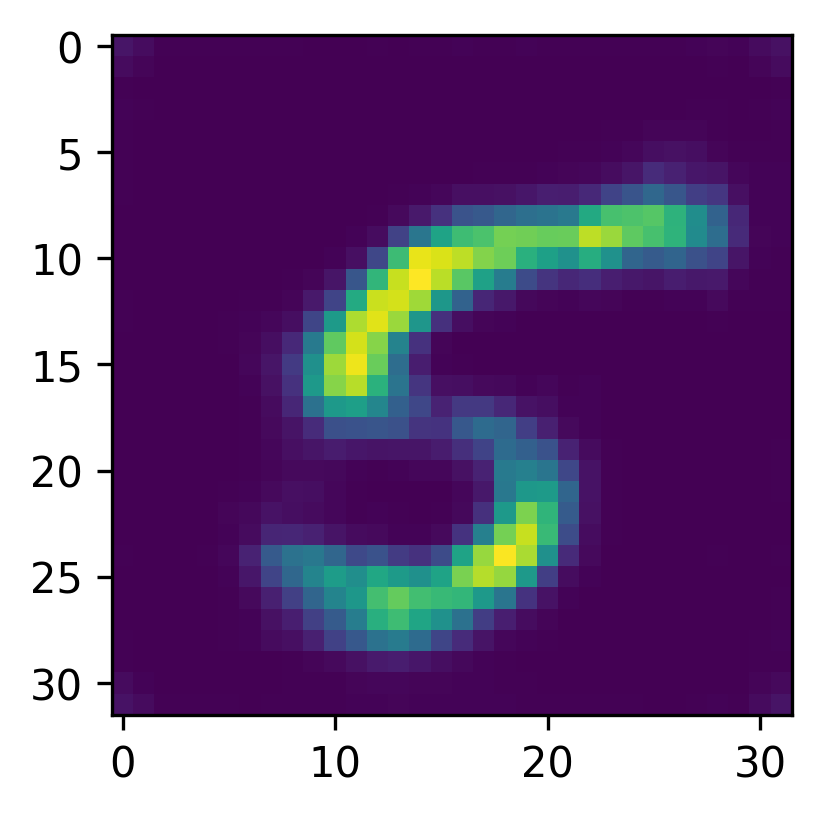}};
  \node[picture format,anchor=north]      (C6) at (B6.south)      {\includegraphics[width=0.10\textwidth]{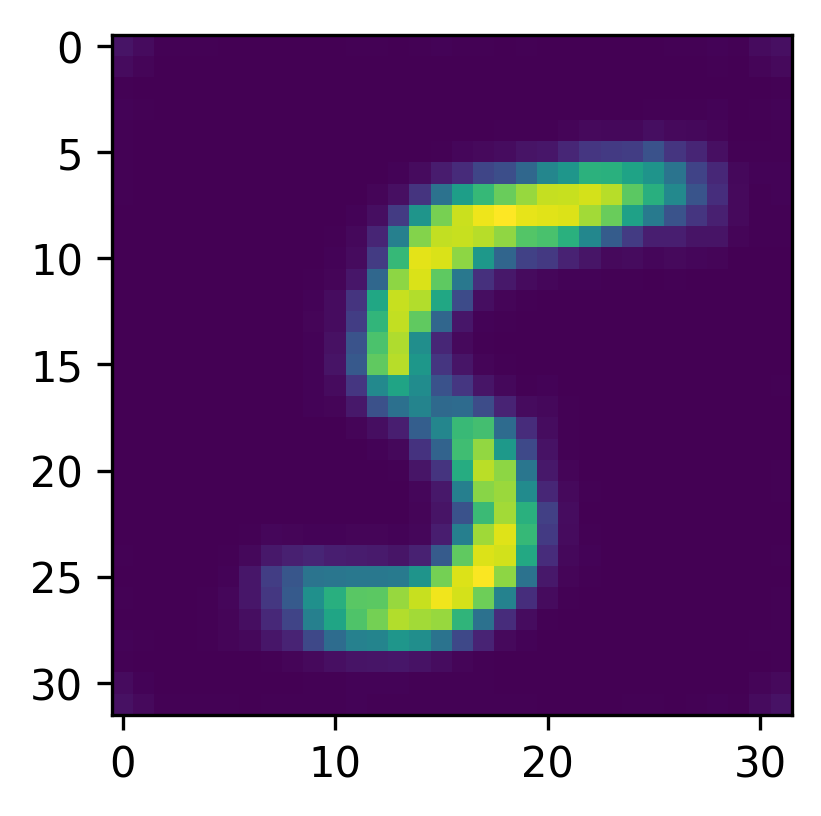}};
  \node[picture format,anchor=north]      (D6) at (C6.south)      {\includegraphics[width=0.10\textwidth]{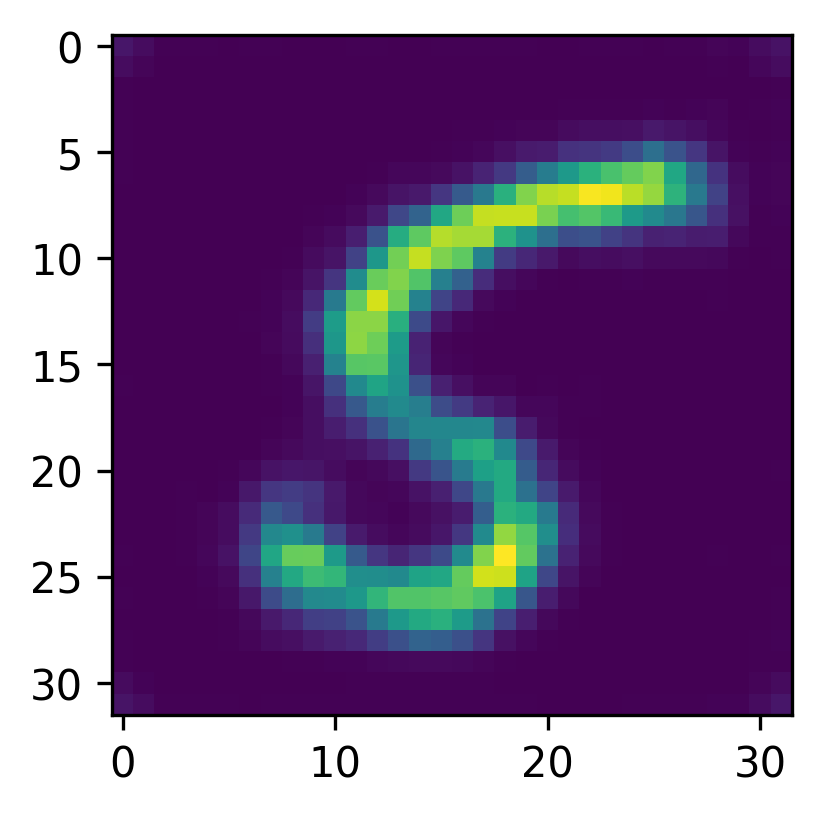}};

  \node[picture format,anchor=north west] (A7) at (A6.north east) {\includegraphics[width=0.10\textwidth]{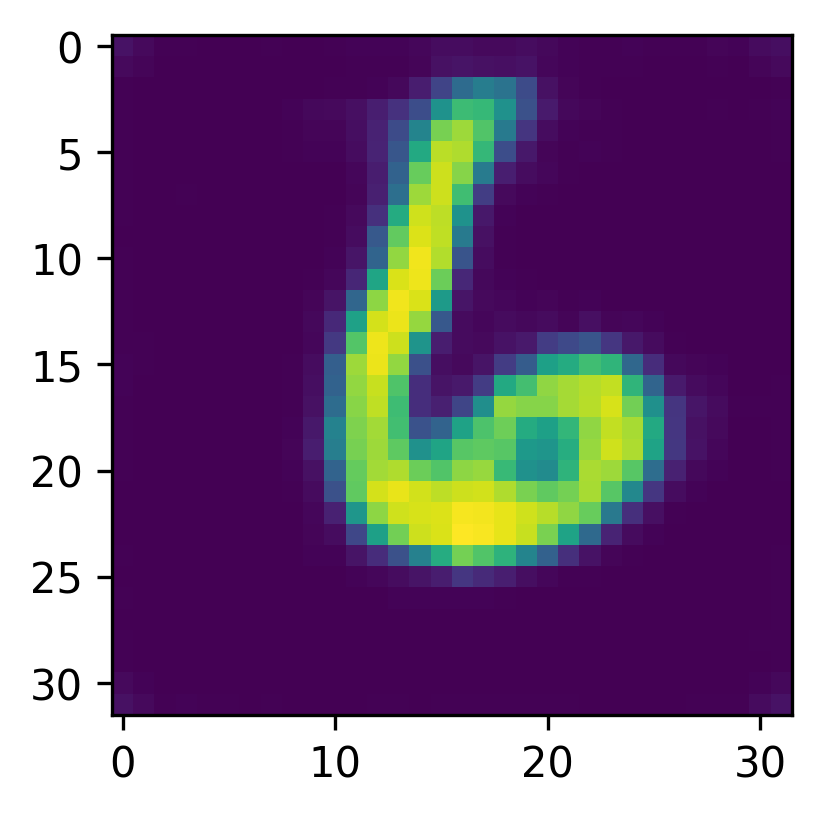}};
  \node[picture format,anchor=north]      (B7) at (A7.south)      {\includegraphics[width=0.10\textwidth]{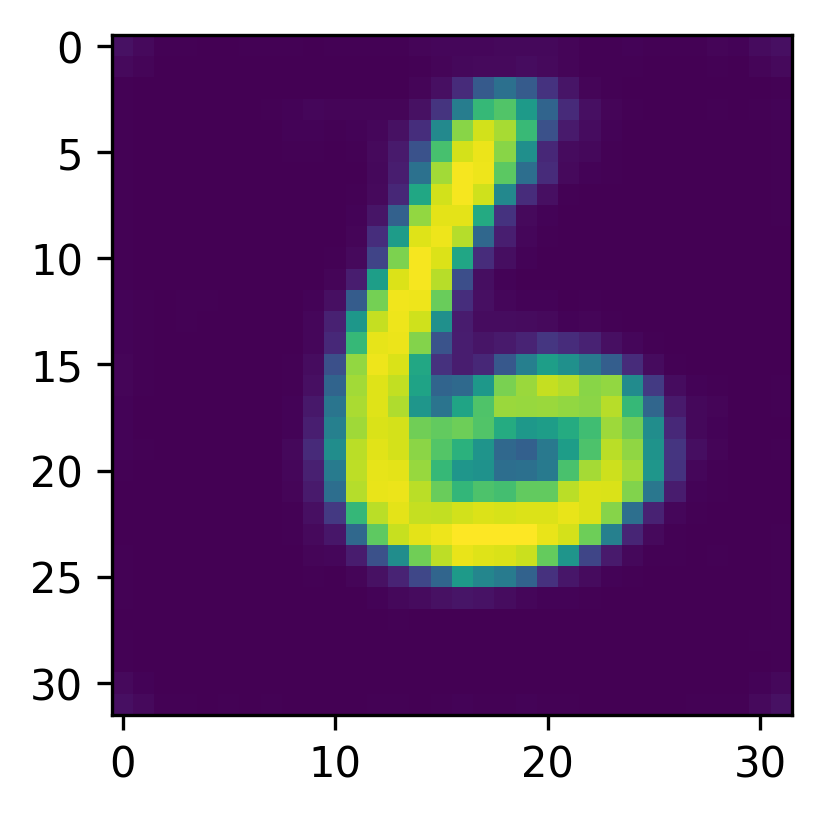}};
  \node[picture format,anchor=north]      (C7) at (B7.south)      {\includegraphics[width=0.10\textwidth]{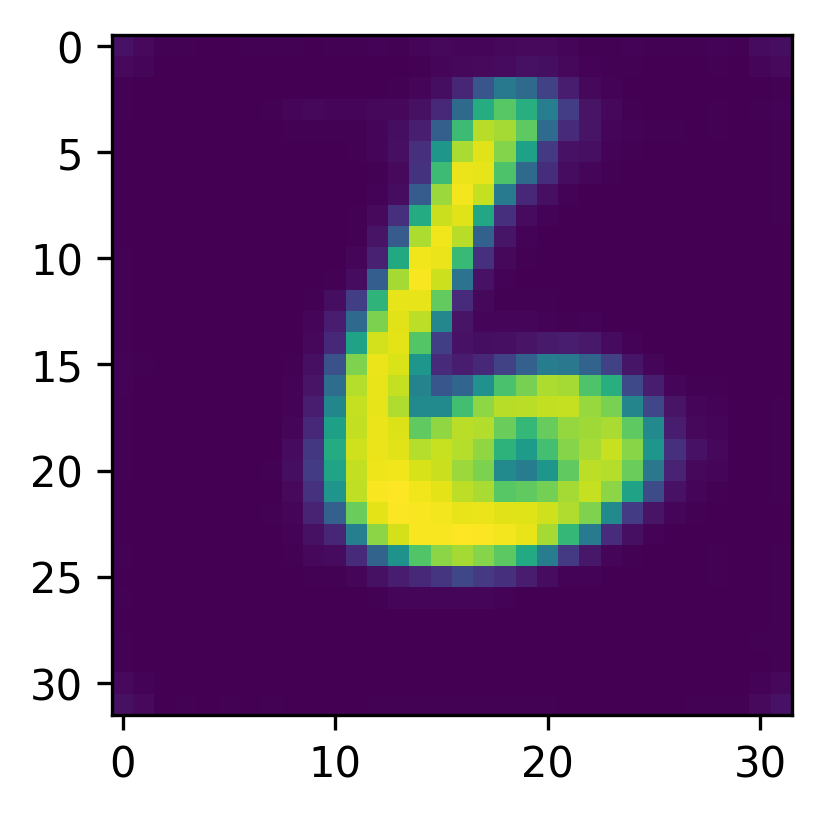}};
  \node[picture format,anchor=north]      (D7) at (C7.south)      {\includegraphics[width=0.10\textwidth]{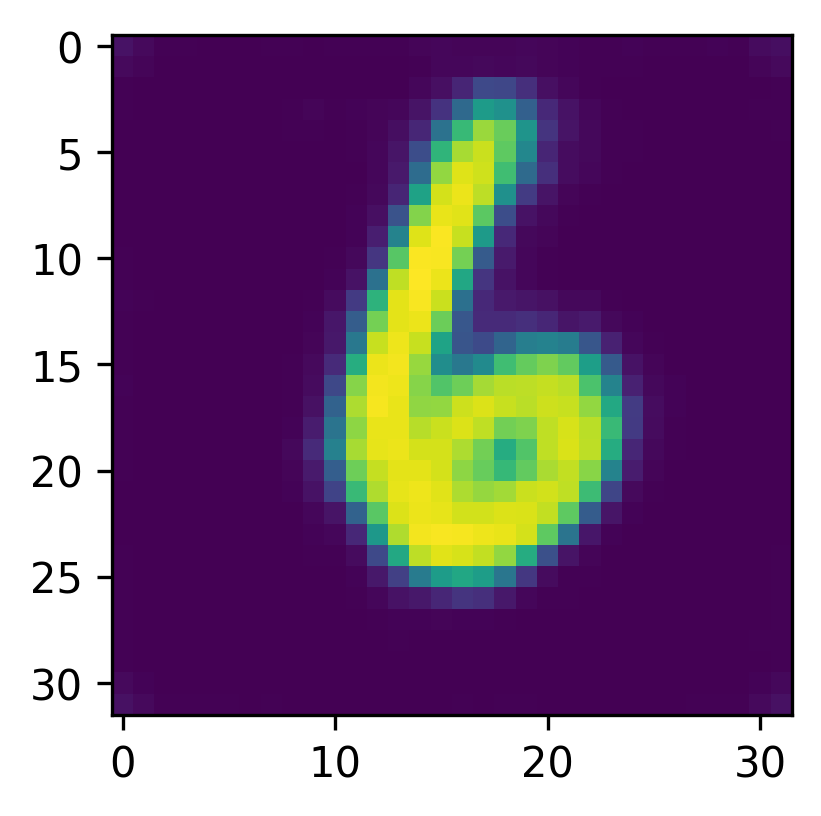}};

  \node[picture format,anchor=north west] (A8) at (A7.north east) {\includegraphics[width=0.10\textwidth]{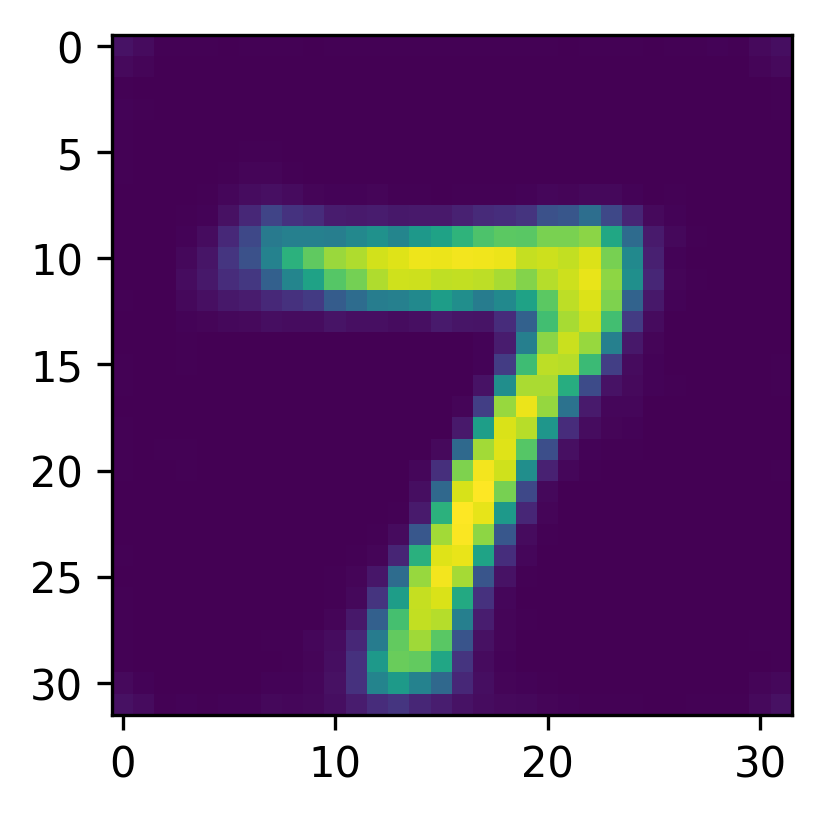}};
  \node[picture format,anchor=north]      (B8) at (A8.south)      {\includegraphics[width=0.10\textwidth]{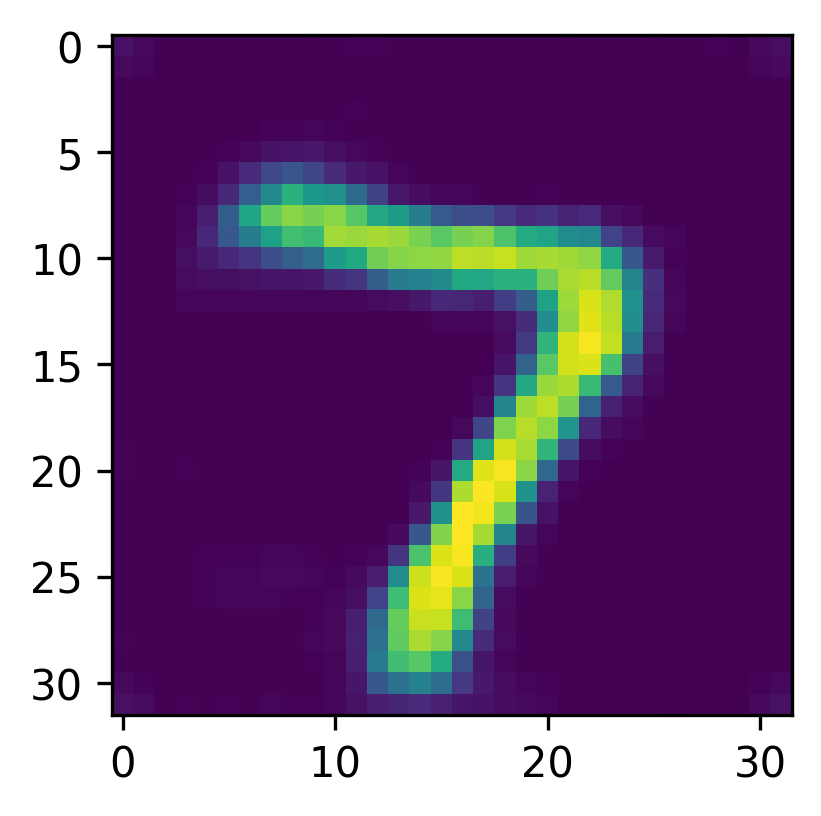}};
  \node[picture format,anchor=north]      (C8) at (B8.south)      {\includegraphics[width=0.10\textwidth]{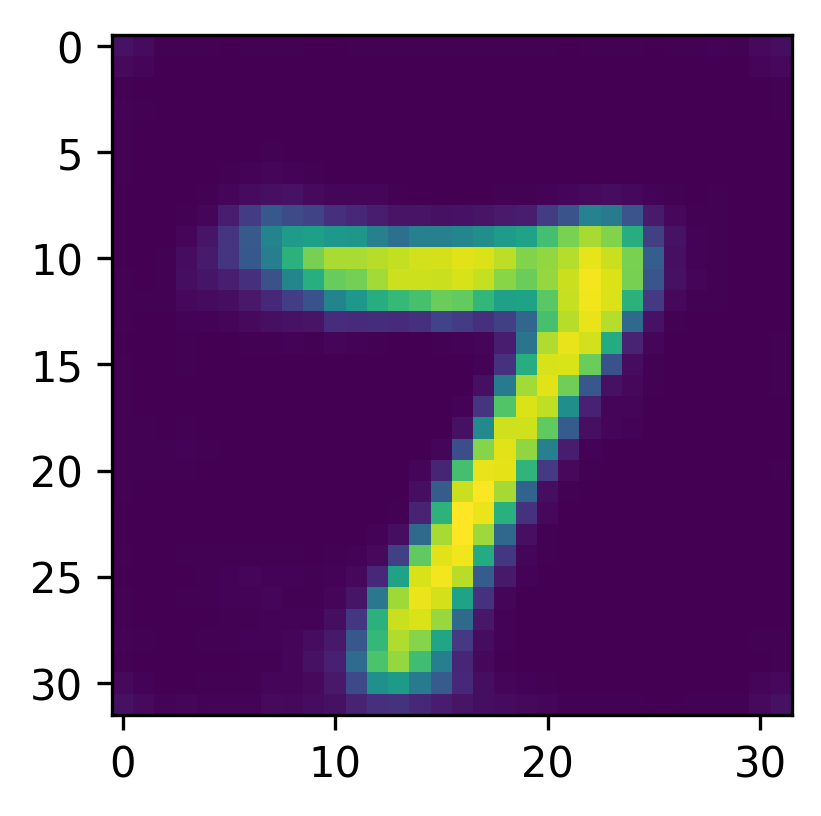}};
  \node[picture format,anchor=north]      (D8) at (C8.south)      {\includegraphics[width=0.10\textwidth]{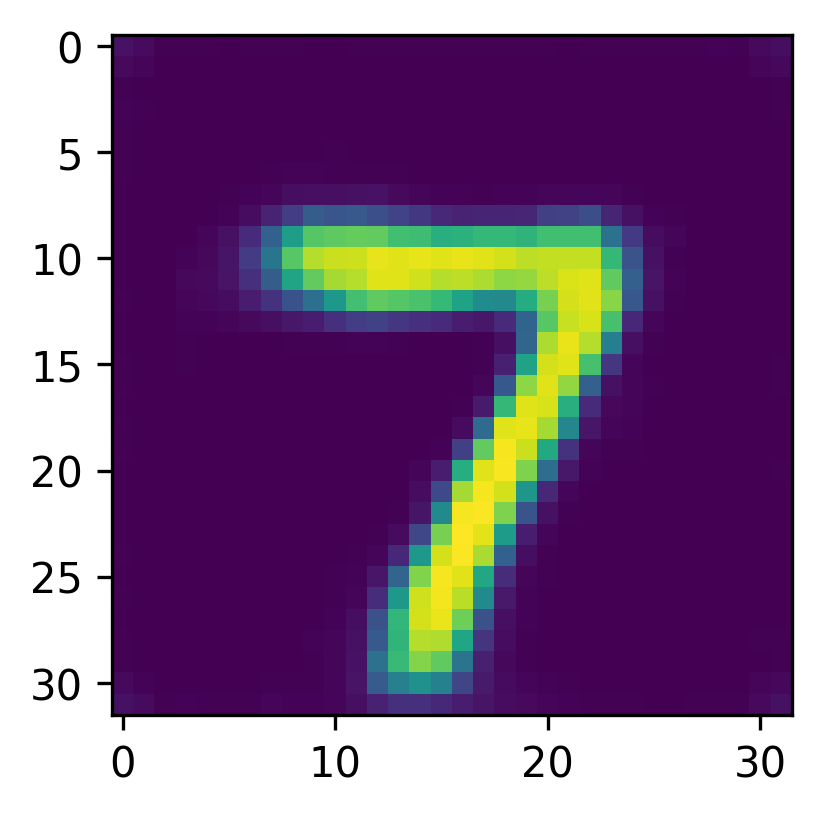}};

  \node[picture format,anchor=north west] (A9) at (A8.north east) {\includegraphics[width=0.10\textwidth]{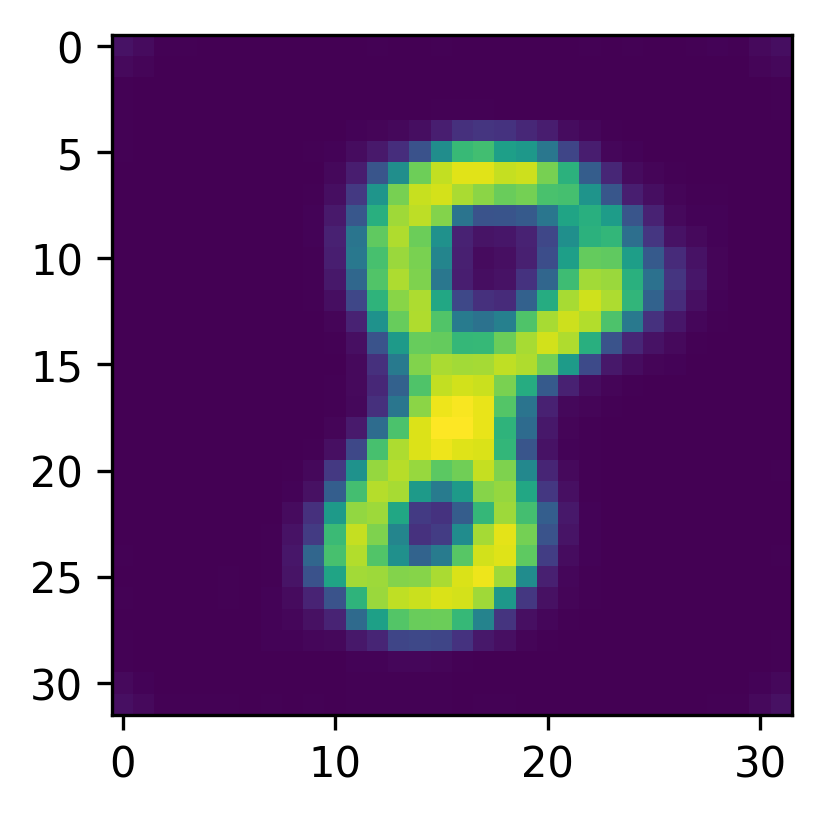}};
  \node[picture format,anchor=north]      (B9) at (A9.south)      {\includegraphics[width=0.10\textwidth]{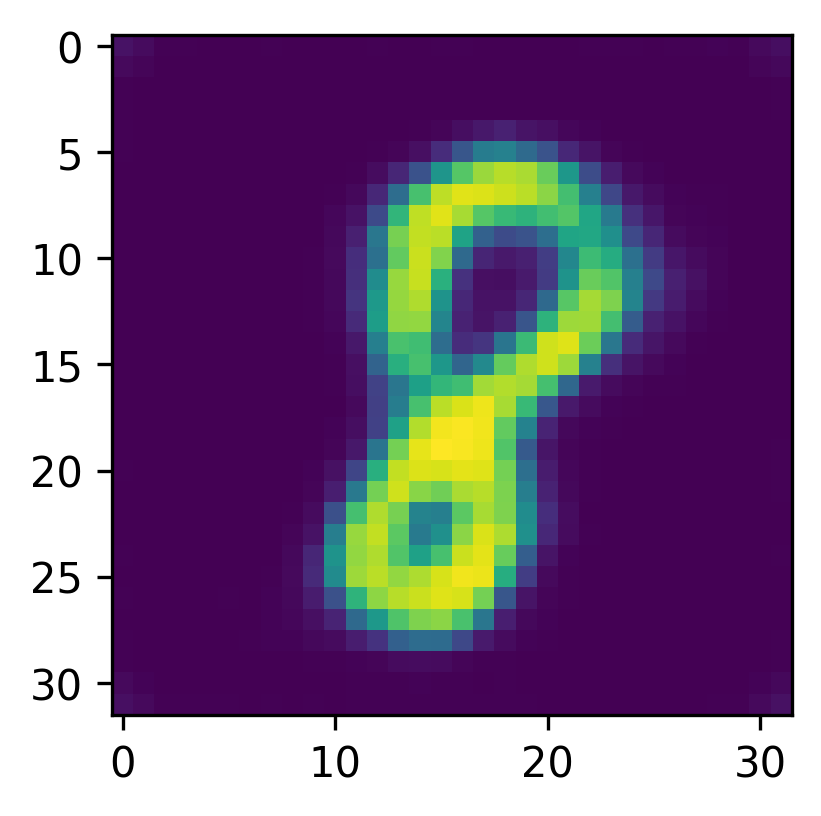}};
  \node[picture format,anchor=north]      (C9) at (B9.south)      {\includegraphics[width=0.10\textwidth]{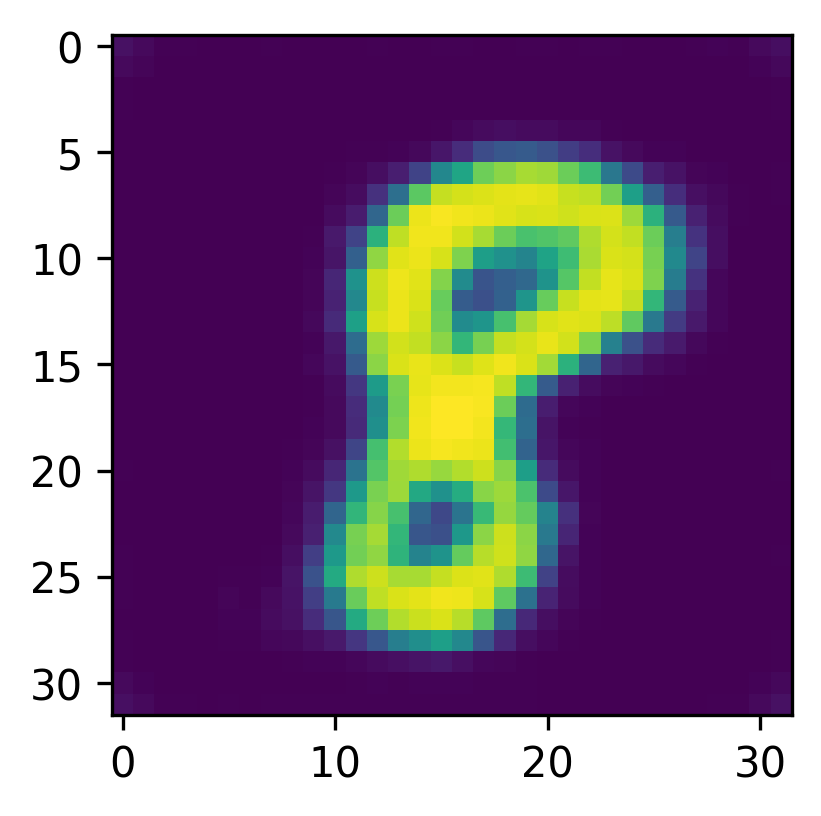}};
  \node[picture format,anchor=north]      (D9) at (C9.south)      {\includegraphics[width=0.10\textwidth]{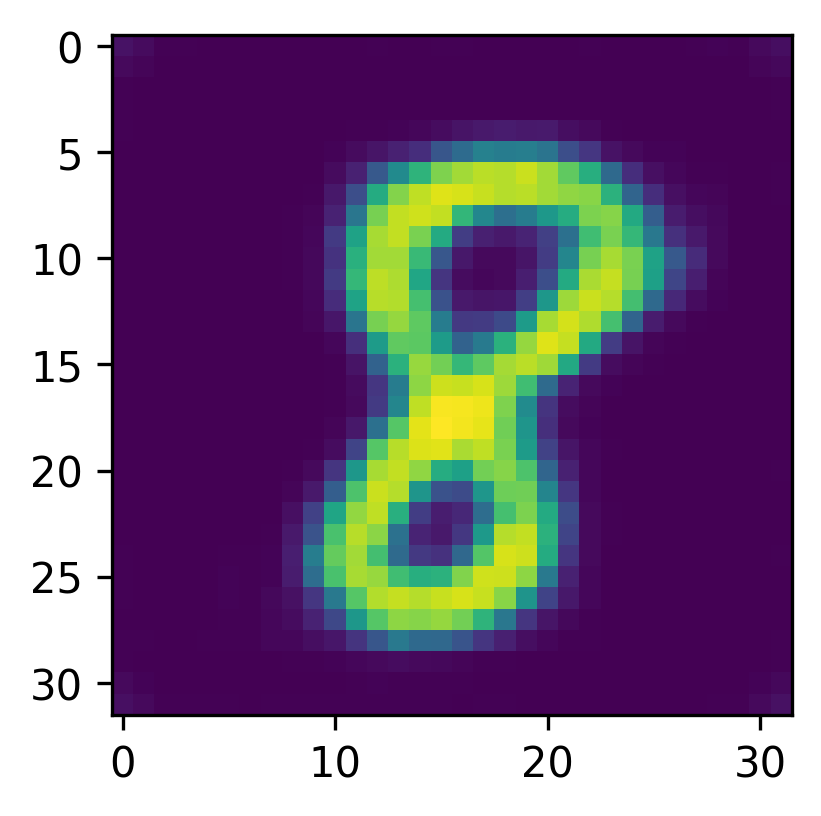}};

  \node[picture format,anchor=north west] (A10) at (A9.north east) {\includegraphics[width=0.10\textwidth]{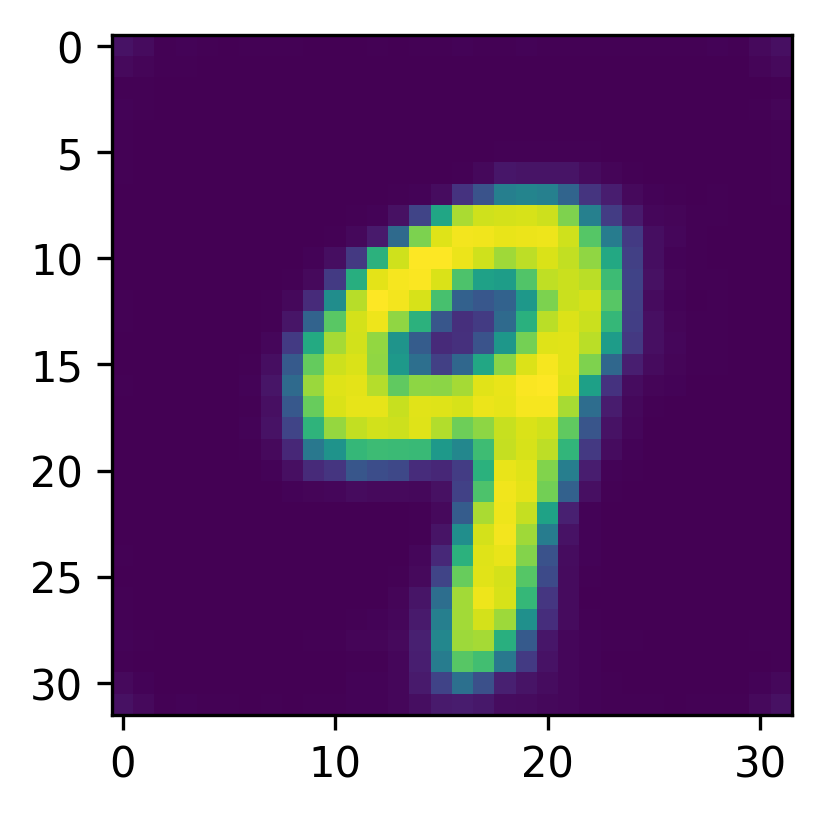}};
  \node[picture format,anchor=north]      (B10) at (A10.south)      {\includegraphics[width=0.10\textwidth]{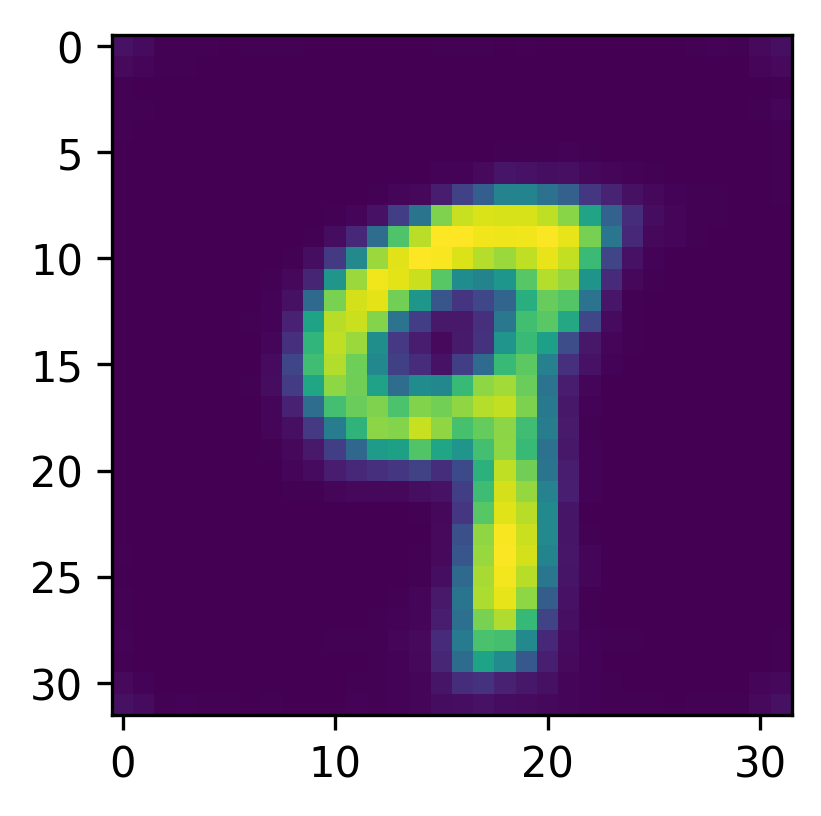}};
  \node[picture format,anchor=north]      (C10) at (B10.south)      {\includegraphics[width=0.10\textwidth]{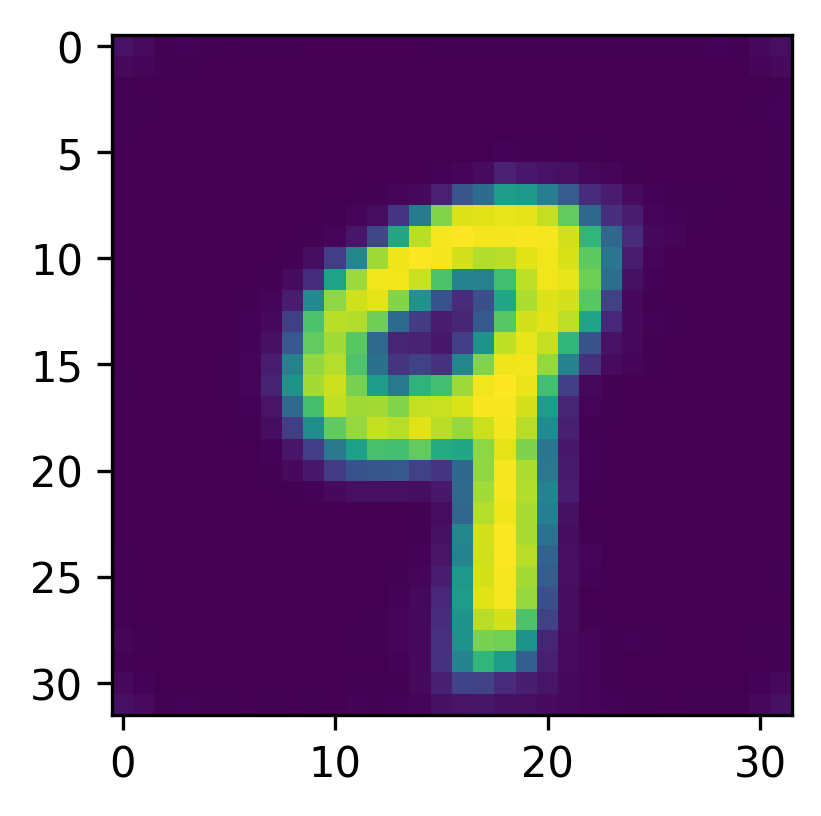}};
  \node[picture format,anchor=north]      (D10) at (C10.south)      {\includegraphics[width=0.10\textwidth]{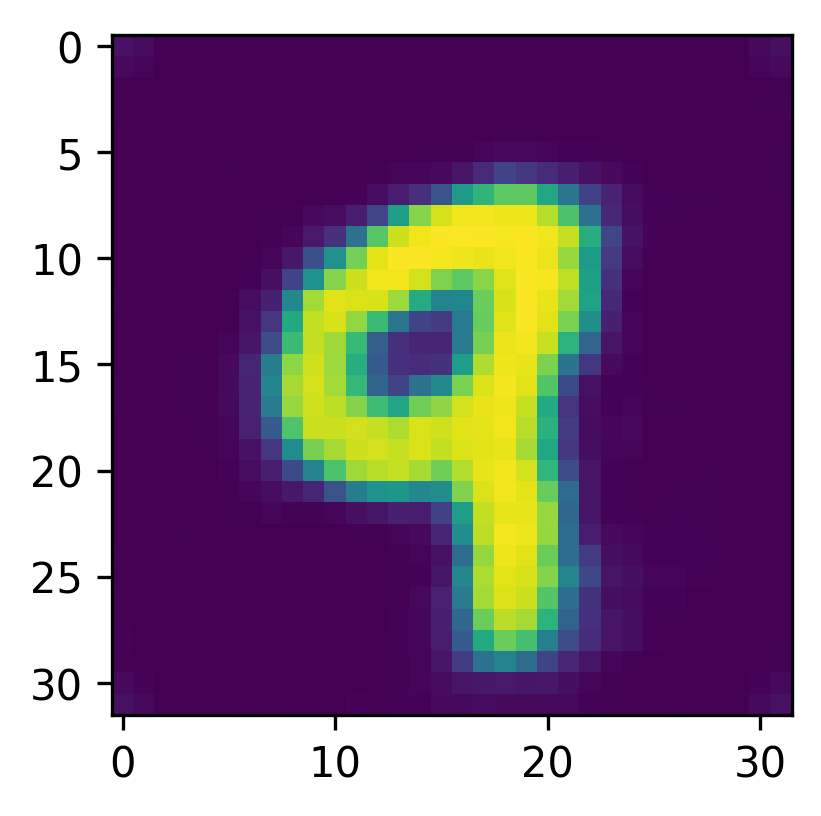}};


  \node[anchor=south] (t1) at (A1.north) {\begin{tabular}{c}
                                            \scriptsize 1\\
                                          \end{tabular}};
  \node[anchor=south] (t2) at (A2.north) {\begin{tabular}{c}
                                            \scriptsize 2 \\
                                          \end{tabular}};
  \node[anchor=south] (t3) at (A3.north) {\begin{tabular}{c}
                                            \scriptsize 3 \\
                                          \end{tabular}};
  \node[anchor=south] (t4) at (A4.north) {\begin{tabular}{c}
                                            \scriptsize 4 \\
                                          \end{tabular}};
  \node[anchor=south] (t5) at (A5.north) {\begin{tabular}{c}
                                            \scriptsize 5 \\
                                          \end{tabular}};
  \node[anchor=south] (t6) at (A6.north) {\begin{tabular}{c}
                                            \scriptsize 6 \\
                                          \end{tabular}};
  \node[anchor=south] (t7) at (A7.north) {\begin{tabular}{c}
                                            \scriptsize 7 \\
                                          \end{tabular}};
  \node[anchor=south] (t8) at (A8.north) {\begin{tabular}{c}
                                            \scriptsize 8 \\
                                          \end{tabular}};
  \node[anchor=south] (t9) at (A9.north) {\begin{tabular}{c}
                                            \scriptsize 9 \\
                                          \end{tabular}};
                                          
  \node[anchor=south] (t10) at (A10.north) {\begin{tabular}{c}
                                            \scriptsize 10 \\
                                          \end{tabular}};                                             

\end{tikzpicture}
\caption{MNIST image generation (recall)}
\label{exp:fig1}
\end{figure}
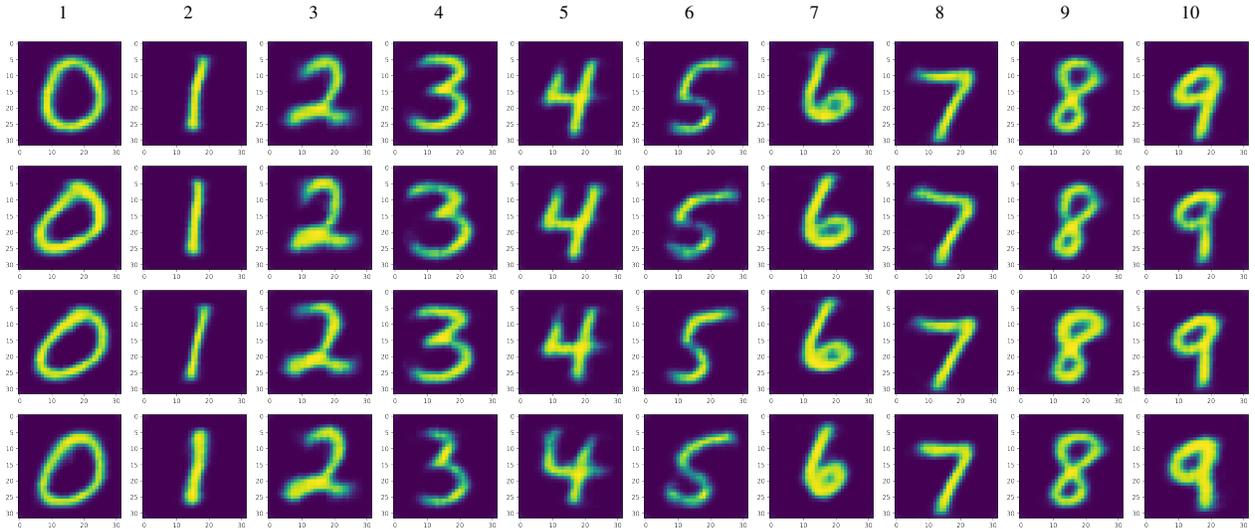

\subsection{Reinforcement Environment}
This section is an experiment in which data is collected in a reinforcement learning environment and screen information is generated for each state. We generated data in the environment with a pre-learned q-learning model, matched screen information with state numbers to create short-term memory, and trained the memory networks model.

\begin{figure}[h]
\centering
\begin{tikzpicture}[picture format/.style={inner sep=0pt,}]
  \node[picture format]                   (A1)               
  {\includegraphics[width=0.12\textwidth]{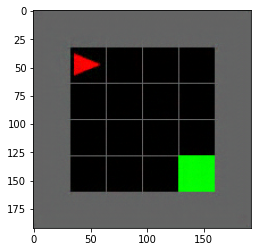}};
  \node[picture format,anchor=north]      (B1) at (A1.south)  {\includegraphics[width=0.12\textwidth]{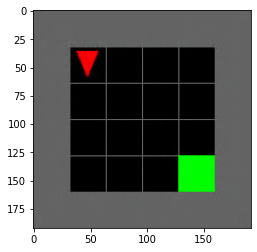}};
  \node[picture format,anchor=north]      (C1) at (B1.south) {\includegraphics[width=0.12\textwidth]{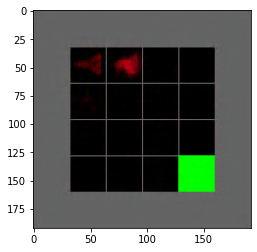}};  
  \node[picture format,anchor=north]      (D1) at (C1.south) {\includegraphics[width=0.12\textwidth]{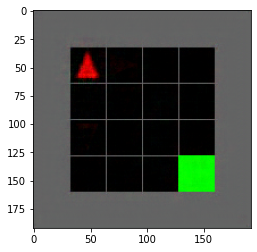}};

  \node[picture format,anchor=north west] (A2) at (A1.north east) {\includegraphics[width=0.12\textwidth]{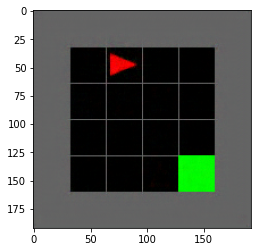}};
  \node[picture format,anchor=north]      (B2) at (A2.south)      {\includegraphics[width=0.12\textwidth]{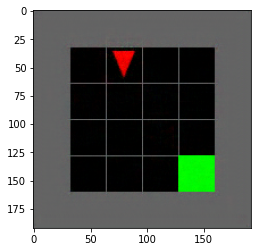}};
  \node[picture format,anchor=north]      (C2) at (B2.south)      {\includegraphics[width=0.12\textwidth]{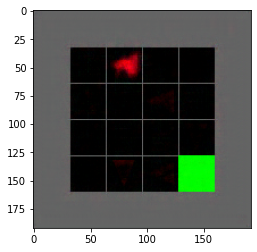}};
  \node[picture format,anchor=north]      (D2) at (C2.south)      {\includegraphics[width=0.12\textwidth]{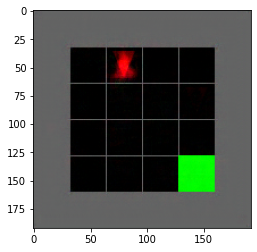}};

  \node[picture format,anchor=north west] (A3) at (A2.north east) {\includegraphics[width=0.12\textwidth]{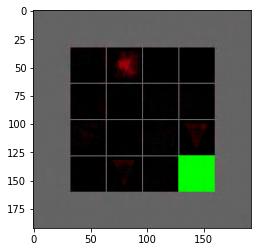}};
  \node[picture format,anchor=north]      (B3) at (A3.south)      {\includegraphics[width=0.12\textwidth]{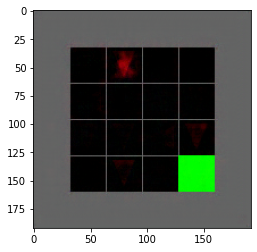}};
  \node[picture format,anchor=north]      (C3) at (B3.south)      {\includegraphics[width=0.12\textwidth]{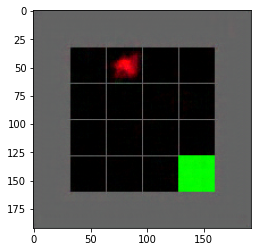}};
  \node[picture format,anchor=north]      (D3) at (C3.south)      {\includegraphics[width=0.12\textwidth]{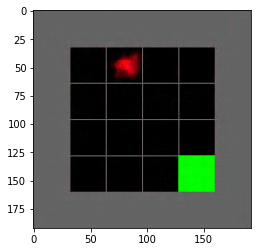}};

  \node[picture format,anchor=north west] (A4) at (A3.north east) {\includegraphics[width=0.12\textwidth]{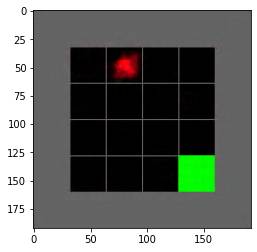}};
  \node[picture format,anchor=north]      (B4) at (A4.south)      {\includegraphics[width=0.12\textwidth]{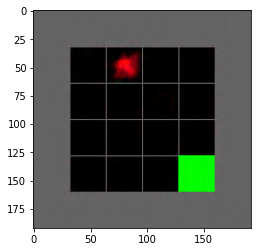}};
  \node[picture format,anchor=north]      (C4) at (B4.south)      {\includegraphics[width=0.12\textwidth]{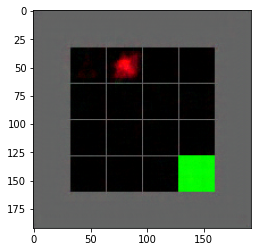}};
  \node[picture format,anchor=north]      (D4) at (C4.south)      {\includegraphics[width=0.12\textwidth]{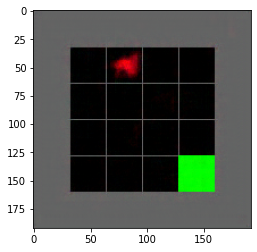}};

  \node[picture format,anchor=north west] (A5) at (A4.north east) {\includegraphics[width=0.12\textwidth]{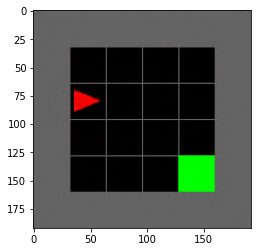}};
  \node[picture format,anchor=north]      (B5) at (A5.south)      {\includegraphics[width=0.12\textwidth]{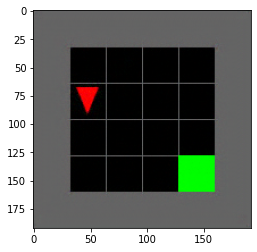}};
  \node[picture format,anchor=north]      (C5) at (B5.south)      {\includegraphics[width=0.12\textwidth]{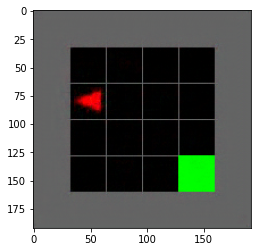}};
  \node[picture format,anchor=north]      (D5) at (C5.south)      {\includegraphics[width=0.12\textwidth]{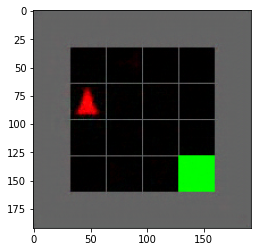}};

  \node[picture format,anchor=north west] (A6) at (A5.north east) {\includegraphics[width=0.12\textwidth]{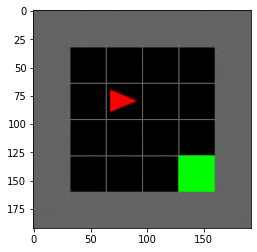}};
  \node[picture format,anchor=north]      (B6) at (A6.south)      {\includegraphics[width=0.12\textwidth]{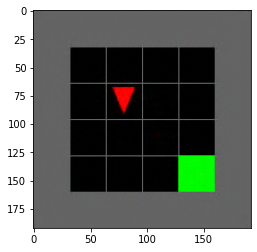}};
  \node[picture format,anchor=north]      (C6) at (B6.south)      {\includegraphics[width=0.12\textwidth]{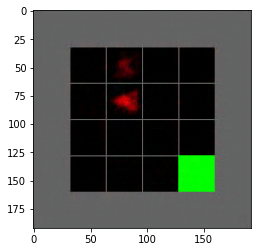}};
  \node[picture format,anchor=north]      (D6) at (C6.south)      {\includegraphics[width=0.12\textwidth]{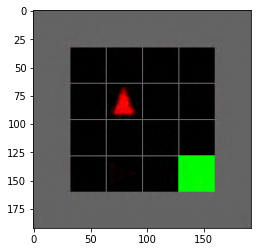}};

  \node[picture format,anchor=north west] (A7) at (A6.north east) {\includegraphics[width=0.12\textwidth]{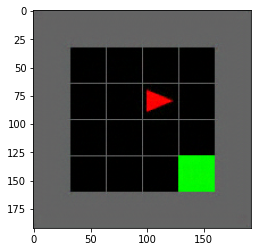}};
  \node[picture format,anchor=north]      (B7) at (A7.south)      {\includegraphics[width=0.12\textwidth]{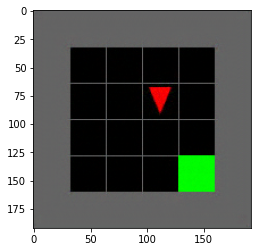}};
  \node[picture format,anchor=north]      (C7) at (B7.south)      {\includegraphics[width=0.12\textwidth]{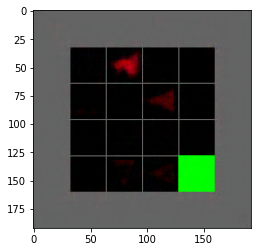}};
  \node[picture format,anchor=north]      (D7) at (C7.south)      {\includegraphics[width=0.12\textwidth]{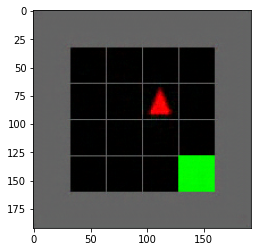}};

  \node[picture format,anchor=north west] (A8) at (A7.north east) {\includegraphics[width=0.12\textwidth]{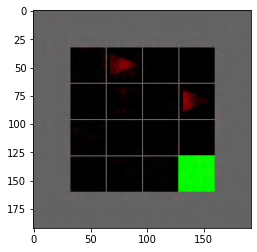}};
  \node[picture format,anchor=north]      (B8) at (A8.south)      {\includegraphics[width=0.12\textwidth]{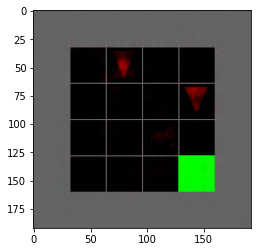}};
  \node[picture format,anchor=north]      (C8) at (B8.south)      {\includegraphics[width=0.12\textwidth]{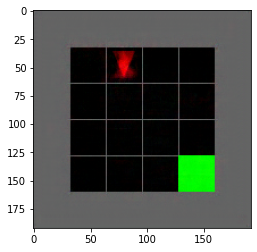}};
  \node[picture format,anchor=north]      (D8) at (C8.south)      {\includegraphics[width=0.12\textwidth]{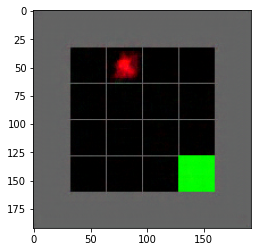}};


  \node[anchor=south] (t1) at (A1.north) {\begin{tabular}{c}
                                            \scriptsize 1\\
                                          \end{tabular}};
  \node[anchor=south] (t2) at (A2.north) {\begin{tabular}{c}
                                            \scriptsize 2 \\
                                          \end{tabular}};
  \node[anchor=south] (t3) at (A3.north) {\begin{tabular}{c}
                                            \scriptsize 3 \\
                                          \end{tabular}};
  \node[anchor=south] (t4) at (A4.north) {\begin{tabular}{c}
                                            \scriptsize 4 \\
                                          \end{tabular}};
  \node[anchor=south] (t5) at (A5.north) {\begin{tabular}{c}
                                            \scriptsize 5 \\
                                          \end{tabular}};
  \node[anchor=south] (t6) at (A6.north) {\begin{tabular}{c}
                                            \scriptsize 6 \\
                                          \end{tabular}};
  \node[anchor=south] (t7) at (A7.north) {\begin{tabular}{c}
                                            \scriptsize 7 \\
                                          \end{tabular}};
  \node[anchor=south] (t8) at (A8.north) {\begin{tabular}{c}
                                            \scriptsize 8 \\
                                          \end{tabular}};

\end{tikzpicture}
\caption{State Screen Recall (from 1 to 8)}
\label{exp:fig1}
\end{figure}
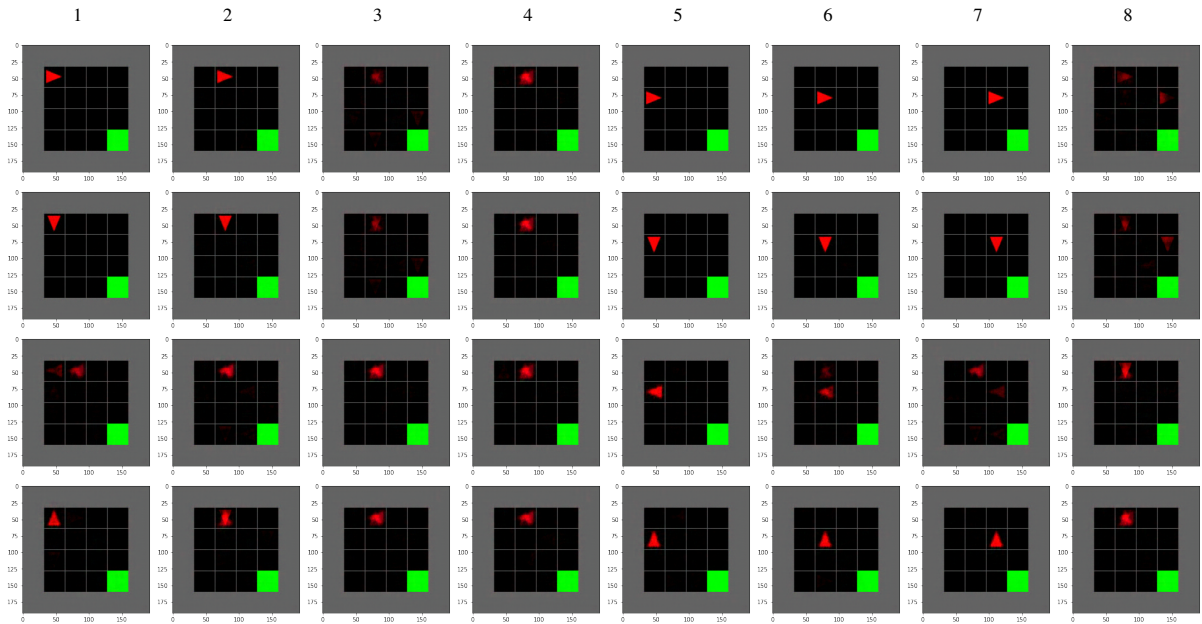

\begin{figure}[h]
\centering
\begin{tikzpicture}[picture format/.style={inner sep=0pt,}]
  \node[picture format]                   (A1)               
  {\includegraphics[width=0.12\textwidth]{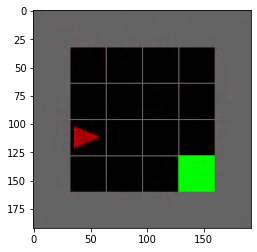}};
  \node[picture format,anchor=north]      (B1) at (A1.south)  {\includegraphics[width=0.12\textwidth]{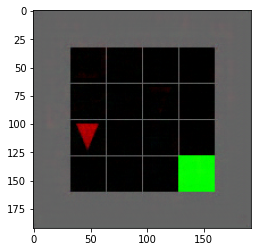}};
  \node[picture format,anchor=north]      (C1) at (B1.south) {\includegraphics[width=0.12\textwidth]{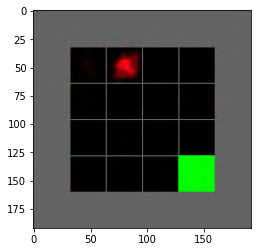}};  
  \node[picture format,anchor=north]      (D1) at (C1.south) {\includegraphics[width=0.12\textwidth]{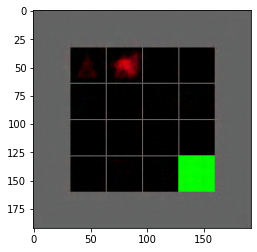}};

  \node[picture format,anchor=north west] (A2) at (A1.north east) {\includegraphics[width=0.12\textwidth]{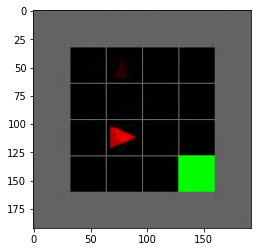}};
  \node[picture format,anchor=north]      (B2) at (A2.south)      {\includegraphics[width=0.12\textwidth]{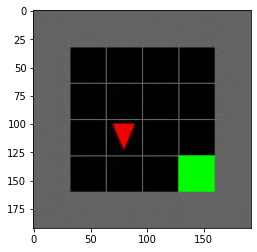}};
  \node[picture format,anchor=north]      (C2) at (B2.south)      {\includegraphics[width=0.12\textwidth]{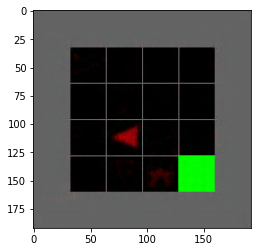}};
  \node[picture format,anchor=north]      (D2) at (C2.south)      {\includegraphics[width=0.12\textwidth]{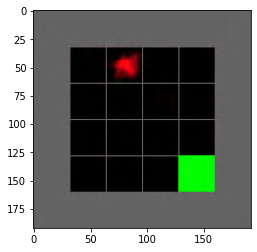}};

  \node[picture format,anchor=north west] (A3) at (A2.north east) {\includegraphics[width=0.12\textwidth]{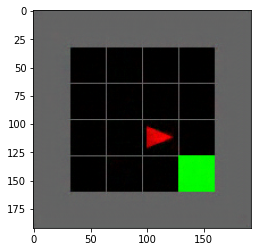}};
  \node[picture format,anchor=north]      (B3) at (A3.south)      {\includegraphics[width=0.12\textwidth]{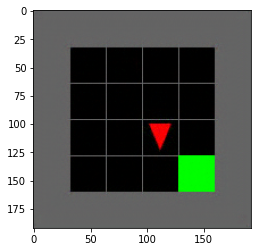}};
  \node[picture format,anchor=north]      (C3) at (B3.south)      {\includegraphics[width=0.12\textwidth]{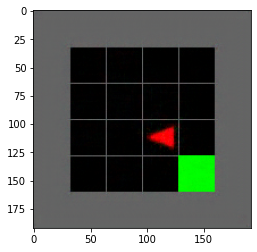}};
  \node[picture format,anchor=north]      (D3) at (C3.south)      {\includegraphics[width=0.12\textwidth]{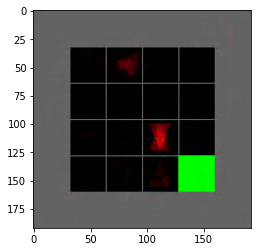}};

  \node[picture format,anchor=north west] (A4) at (A3.north east) {\includegraphics[width=0.12\textwidth]{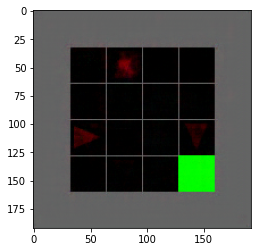}};
  \node[picture format,anchor=north]      (B4) at (A4.south)      {\includegraphics[width=0.12\textwidth]{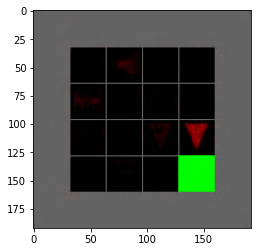}};
  \node[picture format,anchor=north]      (C4) at (B4.south)      {\includegraphics[width=0.12\textwidth]{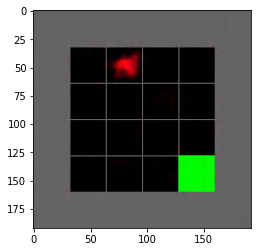}};
  \node[picture format,anchor=north]      (D4) at (C4.south)      {\includegraphics[width=0.12\textwidth]{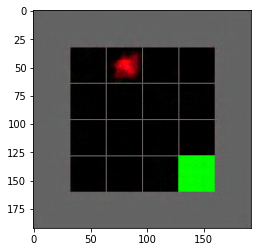}};

  \node[picture format,anchor=north west] (A5) at (A4.north east) {\includegraphics[width=0.12\textwidth]{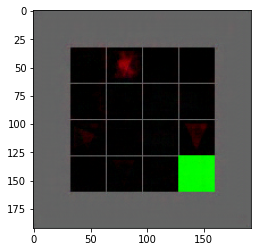}};
  \node[picture format,anchor=north]      (B5) at (A5.south)      {\includegraphics[width=0.12\textwidth]{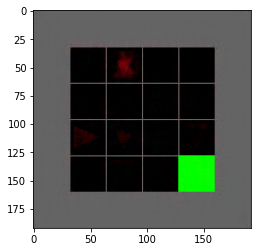}};
  \node[picture format,anchor=north]      (C5) at (B5.south)      {\includegraphics[width=0.12\textwidth]{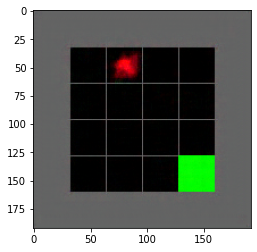}};
  \node[picture format,anchor=north]      (D5) at (C5.south)      {\includegraphics[width=0.12\textwidth]{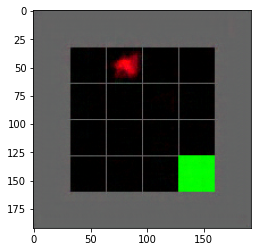}};

  \node[picture format,anchor=north west] (A6) at (A5.north east) {\includegraphics[width=0.12\textwidth]{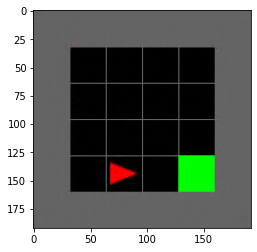}};
  \node[picture format,anchor=north]      (B6) at (A6.south)      {\includegraphics[width=0.12\textwidth]{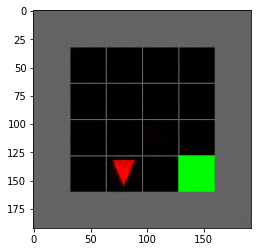}};
  \node[picture format,anchor=north]      (C6) at (B6.south)      {\includegraphics[width=0.12\textwidth]{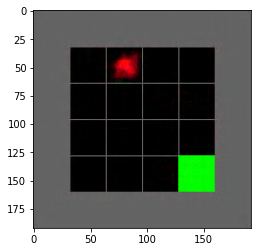}};
  \node[picture format,anchor=north]      (D6) at (C6.south)      {\includegraphics[width=0.12\textwidth]{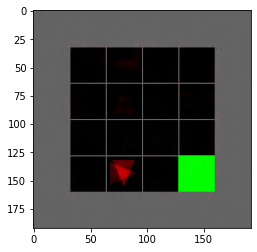}};

  \node[picture format,anchor=north west] (A7) at (A6.north east) {\includegraphics[width=0.12\textwidth]{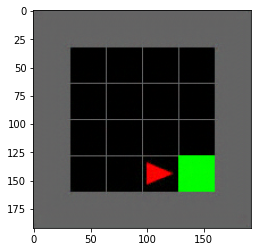}};
  \node[picture format,anchor=north]      (B7) at (A7.south)      {\includegraphics[width=0.12\textwidth]{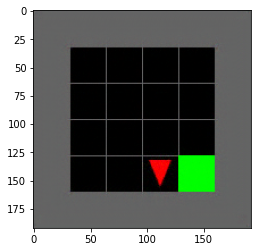}};
  \node[picture format,anchor=north]      (C7) at (B7.south)      {\includegraphics[width=0.12\textwidth]{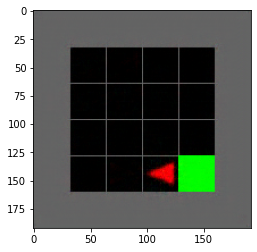}};
  \node[picture format,anchor=north]      (D7) at (C7.south)      {\includegraphics[width=0.12\textwidth]{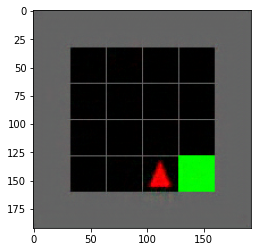}};

  \node[picture format,anchor=north west] (A8) at (A7.north east) {\includegraphics[width=0.12\textwidth]{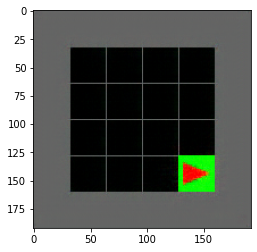}};
  \node[picture format,anchor=north]      (B8) at (A8.south)      {\includegraphics[width=0.12\textwidth]{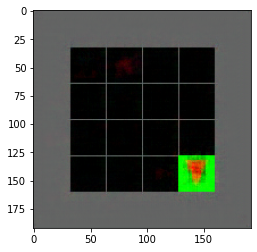}};
  \node[picture format,anchor=north]      (C8) at (B8.south)      {\includegraphics[width=0.12\textwidth]{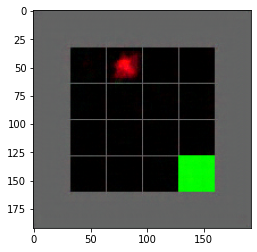}};
  \node[picture format,anchor=north]      (D8) at (C8.south)      {\includegraphics[width=0.12\textwidth]{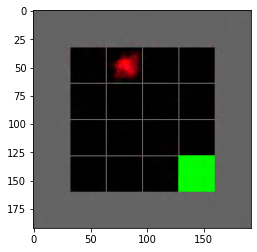}};


  \node[anchor=south] (t1) at (A1.north) {\begin{tabular}{c}
                                            \scriptsize 9\\
                                          \end{tabular}};
  \node[anchor=south] (t2) at (A2.north) {\begin{tabular}{c}
                                            \scriptsize 10 \\
                                          \end{tabular}};
  \node[anchor=south] (t3) at (A3.north) {\begin{tabular}{c}
                                            \scriptsize 11 \\
                                          \end{tabular}};
  \node[anchor=south] (t4) at (A4.north) {\begin{tabular}{c}
                                            \scriptsize 12 \\
                                          \end{tabular}};
  \node[anchor=south] (t5) at (A5.north) {\begin{tabular}{c}
                                            \scriptsize 13 \\
                                          \end{tabular}};
  \node[anchor=south] (t6) at (A6.north) {\begin{tabular}{c}
                                            \scriptsize 14 \\
                                          \end{tabular}};
  \node[anchor=south] (t7) at (A7.north) {\begin{tabular}{c}
                                            \scriptsize 15 \\
                                          \end{tabular}};
  \node[anchor=south] (t8) at (A8.north) {\begin{tabular}{c}
                                            \scriptsize 16 \\
                                          \end{tabular}};

\end{tikzpicture}
\caption{State Screen Recall (from 9 to 16)}
\label{exp:fig1}
\end{figure}
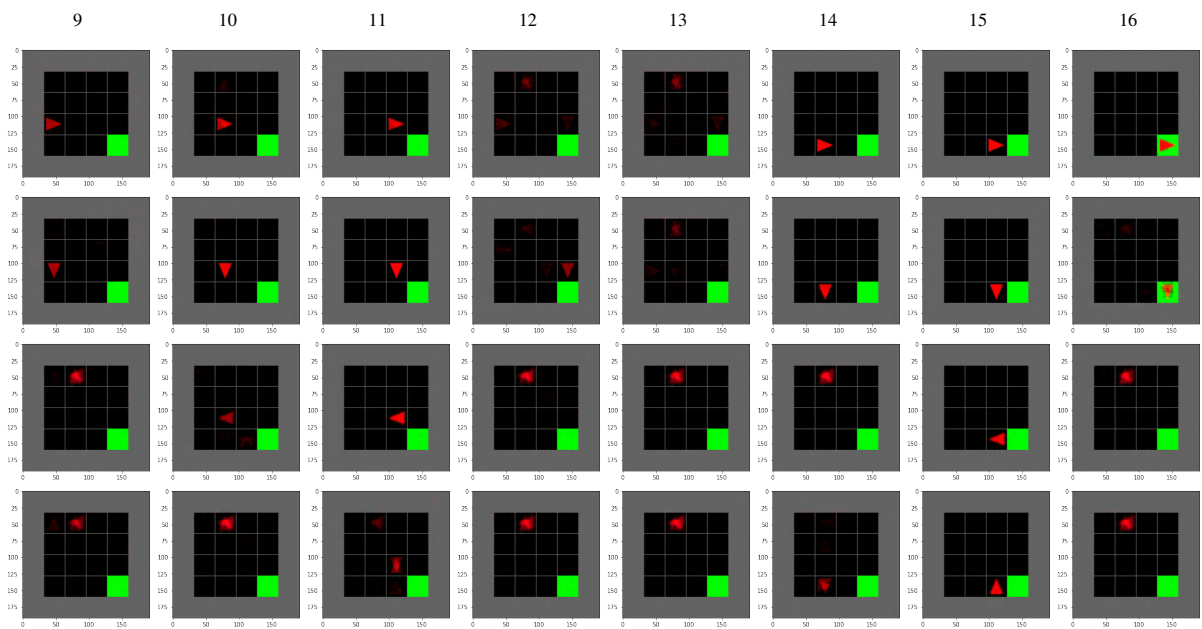
The horizontal is "position" and the vertical is "direction".

Since this result was learned with the image generated by the pre-learned q-learning model, the state information of the shortest path was well learned.
On the other hand, screens in other states did not obtain data and did not learn well.

\section{Conclusion}
We proposed a network model that stores any objects as distributions. This network can generate inputs for deductive association networks.
This paper is part of a series; In the next paper, we introduce Imagine Networks.


\bibliographystyle{splncs04}  
\bibliography{main}


\end{document}